\documentclass[10pt]{article} 
\usepackage[accepted]{tmlr}

\usepackage{amsmath,amsfonts,bm}

\def\eqref#1{equation~\ref{#1}}

\def\1{\bm{1}}

\DeclareMathAlphabet{\mathsfit}{\encodingdefault}{\sfdefault}{m}{sl}
\SetMathAlphabet{\mathsfit}{bold}{\encodingdefault}{\sfdefault}{bx}{n}

\DeclareMathOperator*{\argmax}{arg\,max}

\usepackage[table]{xcolor}
\usepackage{hyperref}
\usepackage{url}

\usepackage[utf8]{inputenc} 
\usepackage[T1]{fontenc}    
\usepackage{microtype}      
\usepackage{inconsolata}    

\usepackage{amsmath, amssymb, amsfonts, amsthm}
\usepackage{bbm}
\usepackage{nicefrac}       
\usepackage{mathabx}

\usepackage{booktabs}       
\usepackage{tabularx}
\usepackage{multirow}
\usepackage{makecell}
\usepackage{arydshln}       
\usepackage{graphicx}
\usepackage{subcaption}
\usepackage{wrapfig}
\usepackage{capt-of}
\usepackage{float}
\usepackage{tabularx}
\usepackage{algorithm}
\usepackage{algpseudocode}
\usepackage{makecell}
\usepackage{booktabs}      
\usepackage{tabularx}      
\usepackage{makecell}      
\usepackage[table]{xcolor} 
\usepackage{subcaption}    
\usepackage[most]{tcolorbox}
\newtcolorbox{ReasoningBox}[1][]{
    colback=gray!5,
    colframe=gray!50,
    fonttitle=\bfseries,
    title={Analysis},
    breakable,
    #1
}
\newtcolorbox{mybox}[1][]{
    title=#1,
    fonttitle=\small,
    fontupper=\small,
    left=2mm,
    right=2mm,
    top=1mm,
    bottom=0mm,
}

\usepackage{xspace}
\usepackage{enumitem}
\setlist{leftmargin=5mm}
\usepackage{textcomp}
\usepackage{pifont}

\definecolor{darkblue}{rgb}{0, 0, 0.5}
\definecolor{correctgreen}{RGB}{0, 153, 0}
\definecolor{wrongred}{RGB}{204, 0, 0}
\hypersetup{
    colorlinks=true,
    citecolor=darkblue,
    linkcolor=darkblue,
    urlcolor=darkblue,
}

\newcommand\ours{\textsc{EmoRouter}\xspace}

\title{EmoRouter: Adaptive Modality Cascading for Multimodal Emotion Recognition}

\author{\name Fangxu Yu \email fxyu@umd.edu \\
      \addr University of Maryland, College Park
      \AND
      \name Lingzhi Yuan \email lingzhiy@umd.edu \\
      \addr University of Maryland, College Park
      \AND
      \name Dehai Min \email dmin10@uic.edu\\
      \addr University of Illinois Chicago
      \AND
      \name Tao Feng  \email taofeng2@illinois.edu\\
      \addr University of Illinois Urbana-Champaign
      \AND
      \name Ge Liu \email geliu@illinois.edu\\
      \addr University of Illinois Urbana-Champaign
      \AND
      \name Furong Huang \email furongh@umd.edu \\
      \addr University of Maryland, College Park
      \AND
      \name Tianyi Zhou\thanks{Corresponding author.} \email tianyi.zhou@mbzuai.ac.ae \\
      \addr Mohamed bin Zayed University of Artificial Intelligence
}

\def\month{12}  
\def\year{2026} 
\def\openreview{\url{https://openreview.net/forum?id=Mfgq0aPP6V}} 

\begin{document}

\maketitle

\begin{abstract}
Multimodal Emotion Recognition (MER) involves detecting the underlying emotion behind each utterance within a conversation, which is essential for emotionally aware human-computer interaction. While omni-modal large language models (Omni-LLMs) demonstrate strong capability for MER by jointly processing modalities including text, audio, and vision, different modality combinations yield different prediction outcomes across utterances: acoustic and visual inputs can help recover annotated labels missed by text-only inference, but can also turn correct predictions into errors. Processing all modalities for every utterance is also computationally expensive. To this end, we introduce \ours{}, a training-free, model-agnostic framework for adaptive modality cascading that selects an exit tier for each utterance along a predefined modality order. The routing is guided by prediction-uncertainty estimates derived from Conformal Prediction.
\ours{} routes each utterance through a sequence of increasing-cost modality levels and exits at the first level when the model is confident in the emotion. To determine when to exit, we propose Conformal Set Contraction, a structural calibration criterion that calibrates the per-tier CP error budgets used for these exit decisions. Comprehensive evaluations on 4 MER benchmarks and 3 Omni-LLMs reveal that \ours{} matches or outperforms the full modalities while reducing inference latency by 67.9--94.5\%. Furthermore, \ours demonstrates high transferability across different models, validating its robustness. Code is available at \href{https://github.com/tianyi-lab/EmoRouter}{https://github.com/tianyi-lab/EmoRouter}.
\end{abstract}

\section{Introduction}

Multimodal emotion recognition (MER) aims to identify the emotional state expressed in an utterance by jointly using textual, acoustic, and visual cues. As a core capability for emotional intelligence, it supports a wide range of high-stakes applications, including human–computer interaction, mental health monitoring, intelligent tutoring, and conversational agents. Human emotions are inherently expressed through multiple complementary sources, including utterance content, vocal prosody, and facial expressions. These heterogeneous modalities are essential for accurate and robust emotion understanding in real-world scenarios.

Traditional MER approaches rely on designing task-specific architectures that fuse features from encoders through graph-based, attention-based, or contrastive learning mechanisms~\citep{yu2024emotion, hu2021mmgcn, hu2022unimse}. While effective, these methods require dataset-specific supervised training and a fixed model architecture, which limits their adaptability and incurs substantial annotation costs. Recently, omni-modal large language models (Omni-LLMs) have emerged as powerful alternatives, capable of jointly processing text, audio, and vision within a unified framework~\citep{zhao2025r1, murzaku2025omnivox, yang2025omni} for MER, delivering competitive performance without task-specific training. However, processing full-modality inputs with Omni-LLMs is expensive due to the massive number of tokens needed to represent images, audio, and video. In addition, the modality combinations under which a model correctly predicts the annotated label can vary across utterances. When text-only predictions are incorrect, acoustic and visual cues may help recover the annotated label. When these predictions are already correct, however, adding modalities incurs computational overhead and may even overturn correct predictions. This trade-off makes fixed modality inputs suboptimal for MER, where modality selection directly affects both predictive performance and computational efficiency.

\begin{wrapfigure}{r}{0.45\textwidth}
\centering
\vspace{-10pt}
\includegraphics[width=0.45\textwidth]{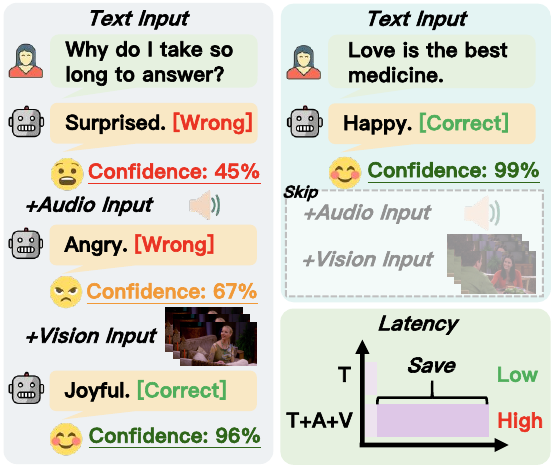}
\caption{Prediction outcomes vary across modality configurations in MER. \textbf{Left:} The model matches the annotated label only with full modalities. \textbf{Right:} Text alone yields a prediction matching the annotation, saving the computational cost of multimodal input. }
\label{fig: intro}
\end{wrapfigure}
To address this, we turn to Conformal Prediction (CP)~\citep{vovk2005algorithmic}, a distribution-free framework that quantifies a model's predictive uncertainty in finite samples. Given a user-specified error tolerance, CP returns a prediction set that is guaranteed to contain the true label with high probability, provided the calibration and test points are exchangeable~\citep{vovk2005algorithmic, angelopoulos2021gentle}. Crucially, the size of this set serves as a direct, interpretable proxy for predictive uncertainty, where a singleton signals a confident prediction, while a larger set reflects ambiguity among labels~\citep{cresswell2024conformal, su2026cp}.
    
Building on the properties of CP, we introduce \ours, a training-free, model-agnostic framework that adaptively selects the exit tier along a predefined modality cascade. Specifically, we construct a three-tier cascade ordered by computational cost, text, text+audio, and text+audio+vision, and exit at the earliest tier where the Omni-LLM is sufficiently certain. For each utterance, we obtain a prediction set from textual cues and use its size as a proxy for uncertainty. Utterances with empty or singleton prediction sets exit with the text-tier top-1 prediction, while those with larger sets are routed to the text-plus-audio tier. Similarly, when the addition of audio still leaves residual uncertainty, visual cues are introduced for emotion recognition. A critical challenge in using CP for MER is selecting appropriate error rates $\alpha$. Therefore, we propose a novel Conformal Set Contraction (CSC) metric that leverages the contraction of prediction sets across tiers to dynamically determine $\alpha$. CSC selects the per-tier budgets $(\alpha_1, \alpha_2)$ that maximize the average set-size shrinkage from one tier to the next on the calibration set, identifying configurations at which each additional modality genuinely reduces uncertainty.

We comprehensively evaluate \ours{} on 4 widely used MER benchmarks and 3 Omni-LLMs. Experimental results indicate that \ours{} matches or outperforms the full modalities while reducing the inference latency by 68.8--94.6\%. Furthermore, \ours{} demonstrates high transferability, as conformal quantile thresholds calibrated on one Omni-LLM transfer directly to different backbones without target-side recalibration. Finally, our analysis demonstrates the effectiveness of CSC compared to a fixed error rate, preserving recognition accuracy while minimizing computational overhead.

\section{Related Work}
\noindent\textbf{Multimodal Emotion Recognition.} Prior MER research has proposed specialized architectures, including graph-based models~\citep{ghosal2019dialoguegcn, ishiwatari2020relation, hu2021mmgcn, shen2021directed, zhang2023dualgats, yu2025beyond} that capture speaker and utterance interactions, transformer- and RNN-based models~\citep{majumder2019dialoguernn, hu2021dialoguecrn, liu2022dialogueein} for temporal context, and contrastive learning methods~\citep{lewis2019bart, song2022supervised, yu2024emotion} that better separate emotion categories. More recent work uses LLMs and multimodal LLM~\citep{lei2023instructerc, zhang2023dialoguellm, wu2024beyond} that predict emotions through instruction tuning and reinforcement learning~\citep{zhao2025r1, lian2025affectgpt, gao2026b, wang2026emotionthinker}. Closest to ours, \citep{chen2025grpo} train a GRPO-tuned LoRA selector for binary routing between text and full modalities, where non-text inputs are converted to captions. \ours{} differs in three ways: it is training-free, runs natively on Omni-LLMs without converting modalities to captions, and transfers across models without recalibration.

\noindent\textbf{Conformal Prediction.}
CP~\citep{vovk2005algorithmic, angelopoulos2021gentle, angelopoulos2024theoretical, gauthier2026backward, liu2026conformal} has been applied to a
range of NLP tasks~\citep{dey2022conformal, maltoudoglou2020bert,
pmlr-v204-ernez23a, su-etal-2024-api, wang2024conu, cresswell2024conformal, silva2025conformal} and, more recently, has emerged as a principled tool for
providing statistically rigorous uncertainty estimates for LLMs: \citep{kumar2023conformal} conformalize softmax scores over multiple-choice answers, and \citep{quach2023conformal} calibrate stopping rules for free-form generation. Closest to our setting, \citep{fisch2020efficient} conformalize prediction cascades, \citep{schuster2021consistent, schuster2022confident} calibrate per-layer early exits in Transformers and LLMs via CP-based stopping rules, and \citep{jitkrittum2023does} show that confidence threshold deferral is provably optimal under broad conditions. In addition, CP variants for non-exchangeable data~\citep{barber2023conformal, gibbs2021adaptive} and for stronger forms of conditional validity~\citep{gibbs2025conformal, cherian2024large} relax the
standard CP assumptions, making CP applicable in increasingly realistic deployment scenarios. The CP-based cascades and deferral rules above operate over model \emph{depth} or \emph{capacity} and take the error level $\alpha$ as user-given; \ours{} instead cascades over the input \emph{modalities} of an Omni-LLM and selects the per-tier error budgets automatically via CSC, contributing CP-driven modality routing with data-driven error-budget allocation for Omni-LLM inference.

\noindent\textbf{LLM Router.} LLM routers aim to achieve a favorable accuracy–efficiency trade-off by routing each query to a suitable LLM, typically based on query embeddings or learned classifiers, such as RouterDC~\citep{chen2024routerdc}, RouteLLM~\citep{ong2024routellm}, GraphPlanner~\citep{feng2026graphplanner} and GraphRouter~\citep{feng2024graphrouter, feng2026llmrouter}, and Hybrid LLM~\citep{ding2024hybrid}. These methods primarily focus on model selection with fixed input modalities. Recent work~\citep{yu2026tsrouter, liu2026efficient} extends routing to modality selection but requires training a dedicated router. In contrast, \ours{} selects modalities based on predictive uncertainty without additional router training.

\section{Method}
\label{sec:method}

\subsection{Motivating Analysis}
\label{ssec:motivation}

We begin with a pilot study on MELD using Qwen3-Omni-30B-A3B-Instruct (abbreviated as Qwen3-Omni-30B in the rest of the paper), which examines the trade-off between computational cost and predictive benefit under fixed modality combinations and yields two key observations that motivate \ours{}.

\begin{figure}[t]
\centering
\includegraphics[width=0.95\textwidth]{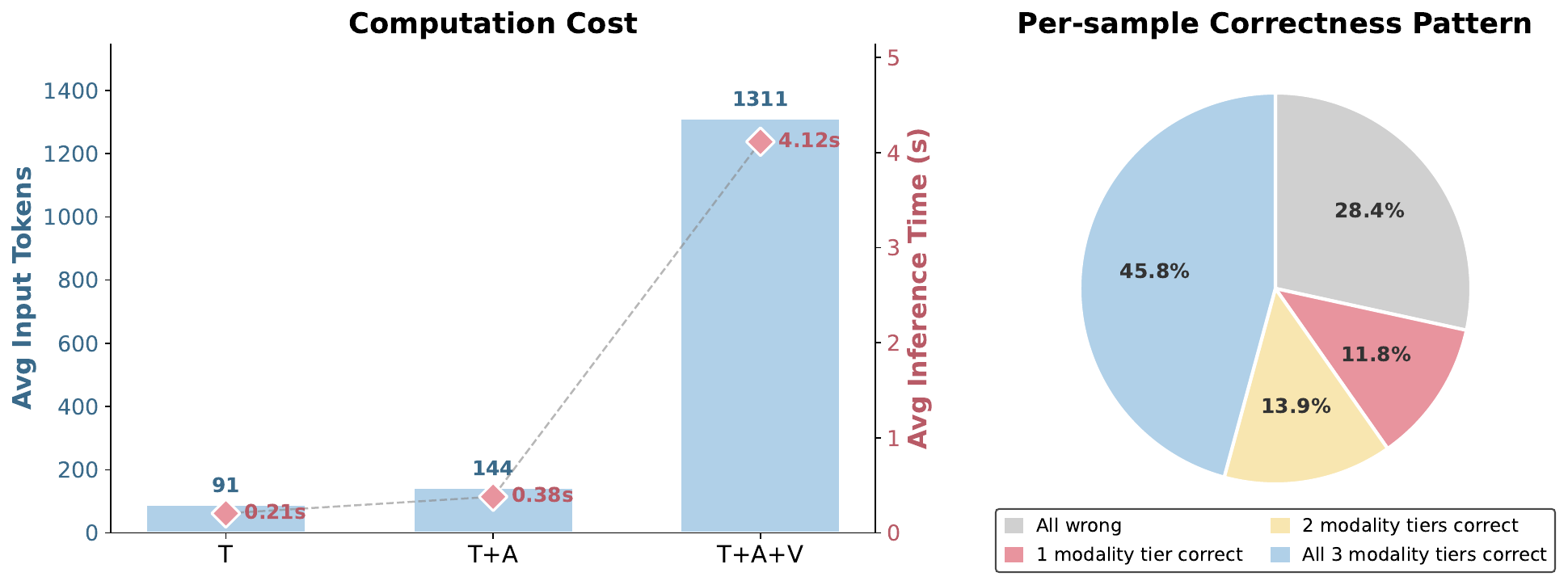}
\caption{Compute cost and modality coverage of Qwen3-Omni-30B on MELD. (a) Average input tokens and inference time per modality tier. (b) Sample distribution across the number of correctly-predicted modality tiers.}
\label{fig:motivation}
\vspace{-10pt}
\end{figure}

\noindent\textbf{More modalities incur greater computational overhead.}
Figure~\ref{fig:motivation} (a) shows that both input-token count and wall-clock latency grow substantially as more modalities are incorporated. Specifically, acoustic cues introduce modest overhead, increasing the average input from 91 to 144 tokens and the average inference time from 0.21\,s to 0.38\,s. In contrast, adding visual cues with full contextual information inflates the input to 1,311 tokens and the inference time to 4.12\,s, making the full-modality tier $19.6\times$ slower than the text-only tier, since video frames must be encoded into far more tokens than the transcript or the audio waveform. Notably, the T+A tier remains $10.8\times$ faster than T+A+V, which makes audio a cost-effective intermediate step before resorting to vision. This observation highlights the need to selectively activate modalities in order to maintain both accuracy and efficiency in emotion recognition, rather than processing all modalities for every utterance by default.

\noindent\textbf{Modality contributions vary across utterances.}
Figure~\ref{fig:motivation} (b) decomposes the test set by the number of modality tiers that predict each utterance correctly relative to its dataset label. We observe that 45.8\% of utterances are predicted correctly by all 3 tiers, i.e., text alone already matches the annotated label, so evaluating additional tiers does not improve label accuracy on these samples. Meanwhile, a non-trivial subset of 25.7\% of utterances can only be correctly predicted by a single modality tier or a specific pair, so their correctness hinges on which modality combination is used, suggesting that judicious modality selection can recover predictions otherwise missed. More modalities do not always help either: in this setting, the full-modality input even underperforms the text-only input (59.0 vs.\ 60.3 wF1; Table~\ref{tab:main_results}), showing that adding acoustic and visual inputs can reduce agreement with the dataset labels. Finally, the remaining 28.4\% of utterances are misclassified at every tier, for which escalating to richer modalities adds cost without recovering the prediction. This diverse pattern shows that the modality configurations yielding correct predictions vary across utterances for the evaluated model, motivating adaptive modality cascading with per-utterance exit decisions.

Together, these findings indicate that an effective router should account for utterance-level differences and select an appropriate exit tier within the predefined modality cascade for each input. This motivates the design of our approach.

\subsection{Problem Formulation}
\label{ssec:problem}

Let $x$ denote an utterance and $y \in \mathcal{Y} = \{y_1, \dots, y_K\}$ its emotion label. A multimodal language model $\mathcal{M}$ can process $x$ at $L$ modality tiers of increasing richness and cost:
\begin{equation}
    \mathcal{T}_1 = \text{T}, \quad
    \mathcal{T}_2 = \text{T{+}A}, \quad
    \mathcal{T}_3 = \text{T{+}A{+}V},
\end{equation}
where T, A, and V denote text, audio, and vision, respectively. Running $x$ through tier $\ell$ incurs a computational cost of $c_\ell(x)$, with $c_1(x) < c_2(x) < c_3(x)$ in practice due to the encoding of audio waveforms and video frames into the token space. Because the cascade evaluates tiers sequentially and the prediction set at each tier is required to decide whether to exit, exiting at tier $\ell$ incurs the cumulative cost $\sum_{j=1}^{\ell} c_j(x)$.

At each tier, $\mathcal{M}$ produces a predictive distribution $\hat{\pi}^{(\ell)}(x) \in \Delta^{K-1}$ over the $K$ emotion classes. 
The goal is to find a per-sample routing function $\ell^*: \mathcal{X} \to \{1, \dots, L\}$ that selects the earliest sufficient tier for each input, minimizing the aggregate computational cost $\sum_i \sum_{\ell=1}^{\ell^*(x_i)} c_\ell(x_i)$ while preserving the classification performance of the full-modality $\mathcal{T}_L$.

\subsection{Preliminaries on Conformal Prediction}
\label{ssec:cp}

Conformal prediction~\citep{vovk2005algorithmic, angelopoulos2021gentle}
is a distribution-free framework for uncertainty quantification. Given a
user-specified error rate $\alpha \in (0,1)$, CP outputs a prediction
set $\mathcal{C}_\alpha(x) \subseteq \mathcal{Y}$ that is guaranteed to
contain the true label with probability at least $1-\alpha$ in finite
samples. The size of $\mathcal{C}_\alpha(x)$ serves as an interpretable
measure of predictive uncertainty: a singleton indicates a confident
prediction, while a larger set reflects ambiguity among
labels~\citep{angelopoulos2021gentle, cresswell2024conformal}.

\noindent\textbf{Nonconformity score.}
The construction begins with a nonconformity score
$s(x, y) \in \mathbb{R}$ that measures how poorly a candidate label $y$
conforms to the model's prediction on $x$. For a classifier producing
predicted probabilities $\hat{\pi}_y(x) \in [0,1]$, a standard choice is
\begin{equation}
    s(x, y) = 1 - \hat{\pi}_y(x),
    \label{eq:score}
\end{equation}
so that a high score indicates that the model assigns low probability to $y$.

\noindent\textbf{Split conformal calibration.}
Let $\mathcal{D}_{\mathrm{cal}} = \{(x_i, y_i)\}_{i=1}^{n}$ be a
calibration set, and assume that
$(x_1,y_1),\dots,(x_n,y_n),(x_{n+1},y_{n+1})$ are exchangeable, where
$(x_{n+1},y_{n+1})$ denotes the test point. The split conformal procedure
turns $s$ into a calibrated set predictor in three steps: (i) compute
the calibration scores $s_i = s(x_i, y_i)$ for $i = 1,\dots,n$; (ii) take
$\hat{q}$ to be the empirical quantile $\lceil (n+1)(1-\alpha) \rceil / n$ 
of $\{s_i\}_{i=1}^{n}$; and (iii) for a new test input $x$, return
\begin{equation}
    \mathcal{C}_\alpha(x) = \bigl\{ y \in \mathcal{Y} : s(x,y) \leq \hat{q} \bigr\},
    \label{eq:predset}
\end{equation}
which, with the score in~\eqref{eq:score}, is equivalent to
$\{y \in \mathcal{Y} : \hat{\pi}_y(x) \geq 1 - \hat{q}\}$.
Under exchangeability, $\mathcal{C}_\alpha(x)$ enjoys the marginal coverage guarantee
\begin{equation}
    \mathbb{P}\bigl(y_{n+1} \in \mathcal{C}_\alpha(x_{n+1})\bigr) \geq 1 - \alpha.
    \label{eq:coverage}
\end{equation}

The cardinality of $\mathcal{C}_\alpha(x)$ is computable from the predicted probabilities alone, requires no labels at test time, and shrinks as the model becomes more confident about $x$. \ours{} exploits this property as the basis for per-tier exit decisions: the cascade exits with the top-1 prediction when $|\mathcal{C}_\alpha(x)|\leq 1$, including an empty-set fallback, and consults additional modalities when the set contains multiple labels.

\begin{figure}[t]   
	
\centering

\includegraphics[width=1.0\textwidth]{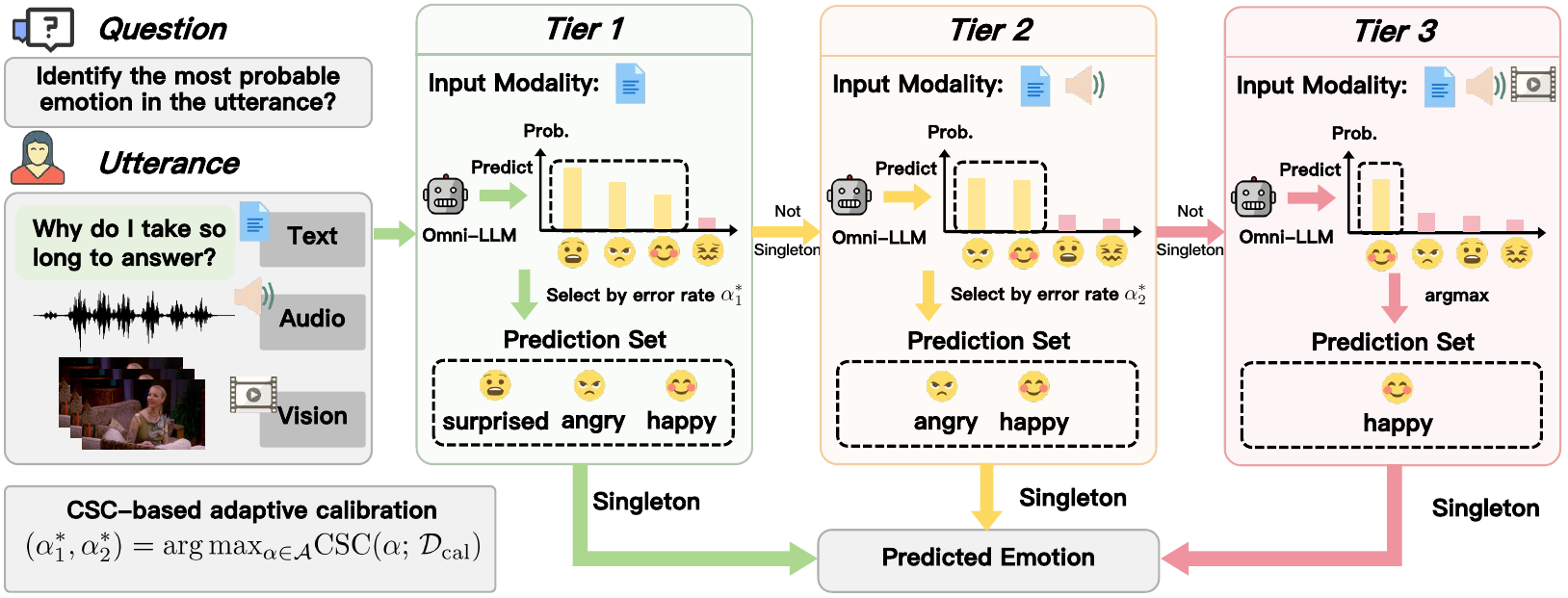}

\caption{Overview of \ours{}. Given an utterance with text, audio, and visual modalities, \ours{} cascades the Omni-LLM through three modality tiers of increasing cost: Tier 1 (T), Tier 2 (T+A), and Tier 3 (T+A+V). At each non-terminal tier, a conformal prediction set is constructed using the CSC-selected error rate $\alpha_{\ell}^{*}$. If the set contains at most one label, the cascade exits and returns the top-1 prediction; otherwise, it escalates to the next tier. The per-tier error budgets are obtained by maximizing Conformal Set Contraction (CSC) on the calibration set.}

\label{fig: main_arch}
\end{figure}
\subsection{\ours Framework}
\label{sec:framework}

\ours consists of three components: (i) per-tier conformal calibration with independent error budgets, (ii) a sequential cascade that exits at the first confident tier, and (iii) a principled procedure for allocating error budgets across tiers.

\noindent\textbf{Per-tier calibration.}
Rather than using a single fixed $\alpha$, we assign each non-terminal tier $\ell \in \{1, \dots, L-1\}$ its own error budget $\alpha_\ell$. For each tier, we compute the nonconformity scores $s_i^{(\ell)} = 1 - \hat{\pi}_{y_i}^{(\ell)}(x_i)$ on $\mathcal{D}_{\mathrm{cal}}$ using the tier-specific predictive distribution $\hat{\pi}^{(\ell)}$, and obtain the threshold
\begin{equation}
    \hat{q}_\ell = \mathrm{Quantile}\Bigl(\{s_i^{(\ell)}\}_{i=1}^{n};\,\tfrac{\lceil (n+1)(1-\alpha_\ell)\rceil}{n}\Bigr),
    \label{eq:qhat_tier}
\end{equation}
i.e., the $\lceil (n+1)(1-\alpha_\ell)\rceil/n$ empirical quantile of the calibration scores. This yields tier-specific prediction sets
$\mathcal{C}_{\alpha_\ell}^{(\ell)}(x) = \{y \in \mathcal{Y} : \hat{\pi}_y^{(\ell)}(x) \geq 1 - \hat{q}_\ell\}$.
The per-tier formulation allows each $\alpha_\ell$ to be tuned independently: a smaller $\alpha_\ell$ yields larger prediction sets at tier $\ell$ and defers more samples to subsequent tiers, while a larger $\alpha_\ell$ shrinks the sets and enables earlier exits. The cardinality $|\mathcal{C}^{(\ell)}_{\alpha_\ell}(x)|$ thus serves as a label-free indicator of tier-$\ell$ confidence and underlies the routing rule introduced next.

\noindent\textbf{Cascade routing.}
At inference time, \ours{} processes each utterance $x$ sequentially through the tiers, using the per-tier prediction sets $\{\mathcal{C}^{(\ell)}_{\alpha_\ell}\}$ as exit signals. For $\ell < L$, the cascade exits at tier $\ell$ if $|\mathcal{C}^{(\ell)}_{\alpha_\ell}(x)| \leq 1$ and returns the top-1 prediction. For a singleton set, this is its unique element. An empty set occurs when no label meets the calibrated probability threshold; the top-1 prediction is used as a fallback in this case. When the set contains multiple labels, the cascade advances to tier $\ell+1$ and incorporates additional modalities. The final tier $\mathcal{T}_L$ serves as an unconditional fallback: $\hat{y} = \arg\max_y \hat{\pi}_y^{(L)}(x)$, regardless of the prediction set size. Formally,
\begin{equation}
\small
    \ell^*(x) = \min\Bigl\{\ell \in \{1, \dots, L\} :
        |\mathcal{C}_{\alpha_\ell}^{(\ell)}(x)| \leq  1
        \;\;\text{or}\;\; \ell = L \Bigr\},
    \label{eq:routing}
\end{equation}
and $\hat{y}(x) = \arg\max_{y} \hat{\pi}_y^{(\ell^*)}(x)$.

The error budgets $\{\alpha_\ell\}$ jointly govern the accuracy--cost trade-off of the cascade: smaller $\alpha_\ell$ produces a larger prediction set at tier $\ell$, encouraging escalation to more expensive tiers, while larger $\alpha_\ell$ admits more early exits, but risks committing prematurely on hard utterances. Because these budgets admit no closed-form optimum and interact across tiers, since $\alpha_1$ determines which utterances reach Tier~2 and thus the population on which $\alpha_2$ operates, we next introduce a data-driven criterion to set them.

\noindent\textbf{Error budget allocation via Conformal Set Contraction.}
To this end, we propose \textbf{Conformal Set Contraction} (CSC), a calibration objective that measures the average rate at which prediction sets contract across consecutive tiers on $\mathcal{D}_{\mathrm{cal}}$. Larger contractions indicate that each additional modality provides genuinely informative content beyond what the previous tier had already resolved, justifying the additional computational cost.
For a candidate budget assignment, let $\hat{q}_\ell$ denote the resulting 
tier-$\ell$ threshold. Initialize $\mathcal{N}_0=\{1,\dots,n\}$ and let $\mathcal{N}_\ell = \{i \in \mathcal{N}_{\ell-1} : 
|\mathcal{C}^{(\ell)}_{\alpha_\ell}(x_i)| > 1\}$ be the calibration samples 
that have not yet exited by tier $\ell$. The average contraction from tier 
$\ell$ to tier $\ell{+}1$ over still-active samples is:
\begin{equation}
    \mathrm{drop}_\ell = \frac{1}{|\mathcal{N}_\ell|} 
    \sum_{i \in \mathcal{N}_\ell} 
    \Bigl(\bigl|\mathcal{C}^{(\ell)}_{\alpha_\ell}(x_i)\bigr| 
         - \bigl|\mathcal{C}^{(\ell+1)}_{\alpha_{\ell+1}}(x_i)\bigr|\Bigr).
\end{equation}
Since the final tier $L$ always exits via $\arg\max$, we adopt the convention $|\mathcal{C}^{(L)}(x_i)| = 1$ when computing $\mathrm{drop}_{L-1}$, and set $\mathrm{drop}_\ell = 0$ when $\mathcal{N}_\ell = \emptyset$, i.e., when no calibration sample remains active at tier $\ell$. CSC is the mean contraction across all non-terminal transitions:
\begin{equation}
    \mathrm{CSC}(\alpha_1,\dots,\alpha_{L-1};\,\mathcal{D}_{\mathrm{cal}})
    = \frac{1}{L-1}\sum_{\ell=1}^{L-1}\mathrm{drop}_\ell.
    \label{eq:CSC}
\end{equation}
Intuitively, CSC balances early resolution and selective escalation:
the first term favors smaller prediction sets at T+A, enabling
cheaper exits, while the final term, with
$|\mathcal{C}^{(L)}|=1$, measures the average residual ambiguity
of samples deferred to T+A+V. Together, these terms encourage
resolving easier samples at lower cost and reserving expensive
inference for ambiguous ones.
We perform a grid search over a candidate grid $\mathcal{A}$ of budget assignments (Appendix~\ref{sec:implementation}), evaluating $\mathrm{CSC}$ on the calibration prediction sets, and select the maximizer:
\begin{equation}
    (\alpha_1^*,\dots,\alpha_{L-1}^*) 
    = \argmax_{\alpha\in\mathcal{A}} 
      \mathrm{CSC}(\alpha;\,\mathcal{D}_{\mathrm{cal}}).
\end{equation}
The selected $(\alpha_1^*,\dots,\alpha_{L-1}^*)$ are then plugged back into the per-tier prediction sets $\{\mathcal{C}^{(\ell)}_{\alpha_\ell^*}\}$, which drive the cascade routing in Equation~\eqref{eq:routing} at inference time. Appendix~\ref{sec: pseudocode} summarizes the full pipeline. Since CSC selects $\alpha^*$ on $\mathcal{D}_{\mathrm{cal}}$ itself, the deployed cascade retains no formal coverage guarantee and uses the prediction-set size purely as a routing signal. 

\section{Experiments}

\subsection{Experimental Setups}
\textbf{Datasets.} To assess \ours across diverse conversational settings and label spaces, we conduct experiments on 4 widely used MER benchmarks that differ in recording condition, speaker interaction, and annotation scheme. IEMOCAP~\citep{busso2008iemocap} contains dyadic conversations performed by 10 professional actors in scripted and improvised scenarios; following prior work~\citep{majumder2019dialoguernn, hu2021mmgcn}, we adopt the 6-class setting of Happy, Sad, Neutral, Angry, Excited, and Frustrated. MELD~\citep{poria2018meld} is a multi-party dataset built from dialogues in the TV series \textit{Friends}, where each utterance is labeled with one of 7 emotions: Anger, Disgust, Fear, Joy, Neutral, Sadness, and Surprise. In contrast to these two conversational datasets, CMU-MOSEI (MOSEI)~\citep{zadeh2018multimodal} and CMU-MOSI (MOSI)~\citep{zadeh2017tensor} consist of single-speaker opinion videos collected from YouTube, where each segment is annotated with a sentiment intensity score in $[-3, 3]$: MOSEI covers more than 1,000 speakers across 250 topics, whereas MOSI focuses on movie reviews from 89 speakers. Together, these benchmarks span acted and spontaneous speech, dyadic and multi-party interaction, and categorical emotion and sentiment polarity, allowing us to stress-test whether \ours adapts its routing to differences in prediction outcomes across datasets and modality configurations. See Table~\ref{tab:dataset_stats} in Appendix~\ref{sec:dataset_details} for dataset statistics and detailed descriptions.

\noindent\textbf{Models.} We include Qwen2.5-Omni-7B, Qwen3-Omni-30B-A3B-Instruct~\citep{xu2025qwen3}, and MiniCPM-o-2.6~\citep{yao2024minicpm}.

\noindent\textbf{Baselines.} 
We compare \ours against five baseline routing strategies. (i) Fixed-modality Routing always selects a single modality. (ii) Random Routing randomly routes utterances across the three tiers. (iii) Self-Consistency Cascade (SCC) exits at Tier 2 when the top-1 predictions of T and T+A agree, and otherwise advances to T+A+V. (iv) Top-1 Probability Routing~\citep{hendrycks2016baseline} uses the highest prediction probability as a proxy for confidence. If this probability exceeds a predefined threshold, the corresponding emotion is output; otherwise, the utterance is routed to the next tier. (v) Entropy Routing routes the utterance to the next tier if the predictive entropy exceeds a predefined threshold, and otherwise outputs the emotion with the highest probability.

\noindent\textbf{Metrics.}
We evaluate each method on effectiveness and efficiency metrics:
(1) Following previous work~\citep{li2023qap, zheng2023facial}, we use \textit{wF1} (Weighted-F1) to measure overall performance.
(2) We evaluate speedup using the Latency Reduction Ratio (LRR), which measures the proportion of computational time saved:
\begin{equation}
    \mathrm{LRR} = 1 - \frac{t_{\text{method}}}{t_{\text{T+A+V}}},
\end{equation}
where $t_{\text{method}}$ and $t_{\text{T+A+V}}$ are the end-to-end wall-clock times for the evaluated method and the full-modality baseline, respectively.
(3) $U_{\text{time}}$ (Time Utility, abbreviated $U_t$ in tables), which captures the improvement in
classification accuracy per unit of wall-clock time, defined in
Equation~\ref{eq:utime} as the ratio of weighted-F1 gain over latency
fraction of a given method relative to full modalities:
\begin{equation}
\label{eq:utime}
U_{\text{time}} \;=\; \frac{\mathrm{wF1} - \mathrm{wF1}_{\text{T}}}
                          {1 - \mathrm{LRR}},
\end{equation}
where $\mathrm{wF1}_{\text{T}}$ is the performance given only textual cues and
$\mathrm{LRR}$ is the latency reduction ratio against T+A+V. A larger
$U_{\text{time}}$ indicates that a method is both more accurate than
T-only and substantially faster than T+A+V.

\begin{table}[t]
\centering
\caption{Main results on four MER benchmarks with three Omni-LLMs, reported as mean$_{\pm\text{std}}$ over three runs. Best mean $U_{t}$ is in \textbf{bold}.}
\label{tab:main_results}
\small
\setlength{\tabcolsep}{3pt}
\renewcommand{\arraystretch}{1.05}
\resizebox{\textwidth}{!}{%
\begin{tabular}{l *{4}{rrr}}
\toprule
& \multicolumn{3}{c}{\textbf{IEMOCAP}} & \multicolumn{3}{c}{\textbf{MELD}} & \multicolumn{3}{c}{\textbf{MOSEI}} & \multicolumn{3}{c}{\textbf{MOSI}} \\
\cmidrule(lr){2-4}\cmidrule(lr){5-7}\cmidrule(lr){8-10}\cmidrule(lr){11-13}
Method & wF1 & LRR & $U_{t}$ & wF1 & LRR & $U_{t}$ & wF1 & LRR & $U_{t}$ & wF1 & LRR & $U_{t}$ \\
\midrule
\rowcolor{gray!12}
\multicolumn{13}{c}{\textit{Qwen2.5-Omni-7B}} \\
T            & 40.7$_{\pm0.5}$ & -- & 0.0$_{\pm0.0}$ & 60.3$_{\pm0.3}$ & -- & 0.0$_{\pm0.0}$ & 58.1$_{\pm1.0}$ & -- & 0.0$_{\pm0.0}$ & 80.5$_{\pm0.4}$ & -- & 0.0$_{\pm0.0}$ \\
T+A+V        & 40.8$_{\pm0.1}$ & 0.0$_{\pm0.0}$ & 0.0$_{\pm0.5}$ & 61.2$_{\pm0.5}$ & 0.0$_{\pm0.0}$ & 1.0$_{\pm0.7}$ & 60.2$_{\pm0.7}$ & 0.0$_{\pm0.0}$ & 2.1$_{\pm0.4}$ & 80.0$_{\pm0.3}$ & 0.0$_{\pm0.0}$ & $-0.5$$_{\pm0.2}$ \\
\hdashline\addlinespace[1pt]
Random       & 43.6$_{\pm0.5}$ & 58.6$_{\pm0.8}$ & 6.9$_{\pm2.3}$ & 61.7$_{\pm0.1}$ & 58.1$_{\pm0.4}$ & 3.4$_{\pm0.7}$ & 59.5$_{\pm1.1}$ & 60.2$_{\pm0.6}$ & 3.5$_{\pm1.0}$ & 81.6$_{\pm0.7}$ & 60.3$_{\pm0.4}$ & 2.7$_{\pm0.9}$ \\
SCC          & 42.9$_{\pm0.3}$ & 62.9$_{\pm0.1}$ & 5.7$_{\pm2.0}$ & 63.2$_{\pm0.4}$ & 63.5$_{\pm0.6}$ & 8.0$_{\pm1.7}$ & 60.5$_{\pm0.8}$ & 81.9$_{\pm0.1}$ & 13.3$_{\pm2.0}$ & 81.7$_{\pm0.5}$ & 78.1$_{\pm1.5}$ & 5.6$_{\pm1.4}$ \\
Top-1@0.7    & 41.1$_{\pm0.1}$ & 71.1$_{\pm0.2}$ & 1.2$_{\pm1.8}$ & 63.3$_{\pm0.3}$ & 70.0$_{\pm0.3}$ & 10.3$_{\pm1.2}$ & 59.8$_{\pm0.6}$ & 79.5$_{\pm0.2}$ & 8.4$_{\pm2.0}$ & 81.2$_{\pm0.1}$ & 80.3$_{\pm0.6}$ & 3.5$_{\pm1.6}$ \\
Top-1@0.9    & 40.9$_{\pm0.0}$ & 31.6$_{\pm0.7}$ & 0.2$_{\pm0.7}$ & 62.5$_{\pm0.4}$ & 32.6$_{\pm1.1}$ & 3.4$_{\pm0.8}$ & 60.2$_{\pm0.8}$ & 44.1$_{\pm0.4}$ & 3.8$_{\pm0.7}$ & 80.4$_{\pm0.2}$ & 50.8$_{\pm1.0}$ & $-0.2$$_{\pm0.3}$ \\
Entropy@0.3  & 40.7$_{\pm0.2}$ & 22.5$_{\pm1.5}$ & $-0.1$$_{\pm0.7}$ & 62.1$_{\pm0.5}$ & 20.8$_{\pm0.6}$ & 2.3$_{\pm1.0}$ & 60.4$_{\pm0.7}$ & 36.2$_{\pm0.7}$ & 3.6$_{\pm0.6}$ & 79.9$_{\pm0.4}$ & 43.7$_{\pm1.2}$ & $-1.1$$_{\pm0.2}$ \\
Entropy@0.7  & 41.3$_{\pm0.0}$ & 60.7$_{\pm0.2}$ & 1.4$_{\pm1.2}$ & 63.4$_{\pm0.1}$ & 54.0$_{\pm0.9}$ & 6.7$_{\pm0.7}$ & 60.1$_{\pm0.7}$ & 78.6$_{\pm0.3}$ & 9.3$_{\pm1.6}$ & 81.4$_{\pm0.2}$ & 76.1$_{\pm1.0}$ & 3.9$_{\pm1.4}$ \\
\rowcolor{blue!6}
\ours        & 46.1$_{\pm0.2}$ & 83.1$_{\pm0.7}$ & \textbf{32.1$_{\pm4.5}$} & 63.6$_{\pm0.5}$ & 87.1$_{\pm0.0}$ & \textbf{25.6$_{\pm5.9}$} & 61.0$_{\pm0.7}$ & 91.9$_{\pm0.0}$ & \textbf{35.1$_{\pm4.5}$} & 82.1$_{\pm0.5}$ & 91.0$_{\pm0.9}$ & \textbf{18.5$_{\pm5.0}$} \\
\addlinespace[2pt]
\midrule
\rowcolor{gray!12}
\multicolumn{13}{c}{\textit{Qwen3-Omni-30B}} \\
T            & 41.4$_{\pm0.5}$ & -- & 0.0$_{\pm0.0}$ & 60.3$_{\pm0.6}$ & -- & 0.0$_{\pm0.0}$ & 66.5$_{\pm0.8}$ & -- & 0.0$_{\pm0.0}$ & 81.5$_{\pm0.9}$ & -- & 0.0$_{\pm0.0}$ \\
T+A+V        & 46.6$_{\pm0.7}$ & 0.0$_{\pm0.0}$ & 5.2$_{\pm1.2}$ & 59.1$_{\pm0.9}$ & 0.0$_{\pm0.0}$ & $-1.2$$_{\pm0.4}$ & 68.3$_{\pm0.3}$ & 0.0$_{\pm0.0}$ & 1.7$_{\pm0.4}$ & 80.7$_{\pm1.5}$ & 0.0$_{\pm0.0}$ & $-0.8$$_{\pm0.5}$ \\
\hdashline\addlinespace[1pt]
Random       & 45.7$_{\pm1.2}$ & 63.6$_{\pm0.8}$ & 12.0$_{\pm4.4}$ & 59.8$_{\pm1.3}$ & 59.6$_{\pm0.8}$ & $-1.2$$_{\pm2.1}$ & 67.7$_{\pm0.5}$ & 52.9$_{\pm0.8}$ & 2.4$_{\pm0.6}$ & 81.7$_{\pm0.6}$ & 61.6$_{\pm0.2}$ & 0.5$_{\pm1.1}$ \\
SCC          & 46.9$_{\pm0.2}$ & 72.3$_{\pm0.6}$ & 20.0$_{\pm1.5}$ & 61.6$_{\pm0.9}$ & 68.3$_{\pm0.5}$ & 4.0$_{\pm1.2}$ & 68.2$_{\pm0.6}$ & 71.5$_{\pm0.9}$ & 6.0$_{\pm0.4}$ & 82.1$_{\pm1.6}$ & 82.1$_{\pm0.6}$ & 3.4$_{\pm3.9}$ \\
Top-1@0.7    & 44.4$_{\pm0.2}$ & 94.3$_{\pm0.3}$ & 53.1$_{\pm2.2}$ & 61.1$_{\pm0.7}$ & 91.8$_{\pm0.1}$ & 7.9$_{\pm2.7}$ & 67.1$_{\pm0.8}$ & 88.5$_{\pm0.2}$ & 4.7$_{\pm0.3}$ & 81.9$_{\pm1.2}$ & 95.3$_{\pm0.2}$ & \textbf{8.7$_{\pm7.0}$} \\
Top-1@0.9    & 45.7$_{\pm0.6}$ & 82.2$_{\pm0.3}$ & 24.8$_{\pm6.1}$ & 61.7$_{\pm0.7}$ & 81.7$_{\pm0.6}$ & 7.4$_{\pm0.4}$ & 67.3$_{\pm0.9}$ & 82.0$_{\pm0.5}$ & 4.1$_{\pm1.1}$ & 82.2$_{\pm1.3}$ & 89.7$_{\pm0.8}$ & 6.7$_{\pm3.7}$ \\
Entropy@0.3  & 46.0$_{\pm0.7}$ & 78.7$_{\pm0.1}$ & 21.7$_{\pm5.2}$ & 61.7$_{\pm0.7}$ & 78.1$_{\pm0.5}$ & 6.5$_{\pm0.4}$ & 67.5$_{\pm0.9}$ & 80.9$_{\pm0.6}$ & 5.1$_{\pm1.0}$ & 82.1$_{\pm1.4}$ & 89.2$_{\pm0.7}$ & 5.4$_{\pm4.2}$ \\
Entropy@0.7  & 43.4$_{\pm0.3}$ & 94.6$_{\pm0.3}$ & 37.5$_{\pm5.1}$ & 60.9$_{\pm0.4}$ & 91.4$_{\pm0.1}$ & 7.2$_{\pm1.6}$ & 66.7$_{\pm0.8}$ & 89.5$_{\pm0.0}$ & 1.9$_{\pm0.5}$ & 81.9$_{\pm1.1}$ & 95.0$_{\pm0.2}$ & 7.5$_{\pm3.3}$ \\
\rowcolor{blue!6}
\ours        & 49.4$_{\pm0.6}$ & 94.2$_{\pm0.8}$ & \textbf{142.5$_{\pm6.1}$} & 62.7$_{\pm0.6}$ & 90.4$_{\pm0.2}$ & \textbf{25.0}$_{\pm1.5}$ & 68.5$_{\pm0.5}$ & 84.4$_{\pm0.3}$ & \textbf{12.7$_{\pm1.9}$} & 81.4$_{\pm1.3}$ & 72.5$_{\pm6.6}$ & $-0.8$$_{\pm1.8}$ \\
\addlinespace[2pt]
\midrule
\rowcolor{gray!12}
\multicolumn{13}{c}{\textit{MiniCPM-o-2.6}} \\
T            & 37.8$_{\pm0.4}$ & -- & 0.0$_{\pm0.0}$ & 59.7$_{\pm0.4}$ & -- & 0.0$_{\pm0.0}$ & 55.9$_{\pm0.7}$ & -- & 0.0$_{\pm0.0}$ & 71.1$_{\pm0.9}$ & -- & 0.0$_{\pm0.0}$ \\
T+A+V        & 40.5$_{\pm0.4}$ & 0.0$_{\pm0.0}$ & 2.8$_{\pm0.6}$ & 60.4$_{\pm0.4}$ & 0.0$_{\pm0.0}$ & 0.7$_{\pm0.8}$ & 57.5$_{\pm1.2}$ & 0.0$_{\pm0.0}$ & 1.7$_{\pm0.5}$ & 74.9$_{\pm1.2}$ & 0.0$_{\pm0.0}$ & 3.8$_{\pm0.6}$ \\
\hdashline\addlinespace[1pt]
Random       & 40.5$_{\pm0.8}$ & 56.7$_{\pm0.9}$ & 6.2$_{\pm1.1}$ & 61.2$_{\pm0.3}$ & 63.6$_{\pm0.2}$ & 4.0$_{\pm1.3}$ & 57.2$_{\pm1.2}$ & 61.3$_{\pm0.7}$ & 3.4$_{\pm1.3}$ & 74.5$_{\pm1.1}$ & 58.0$_{\pm0.2}$ & 8.1$_{\pm0.5}$ \\
SCC          & 40.9$_{\pm0.5}$ & 66.8$_{\pm0.2}$ & 9.4$_{\pm0.5}$ & 63.3$_{\pm0.3}$ & 72.3$_{\pm0.9}$ & 13.0$_{\pm0.8}$ & 58.8$_{\pm1.3}$ & 73.4$_{\pm0.6}$ & 11.0$_{\pm2.3}$ & 77.5$_{\pm1.1}$ & 56.8$_{\pm0.9}$ & 14.9$_{\pm0.6}$ \\
Top-1@0.7    & 39.7$_{\pm0.4}$ & 84.3$_{\pm0.7}$ & 12.5$_{\pm0.3}$ & 62.0$_{\pm0.1}$ & 82.9$_{\pm0.4}$ & 13.4$_{\pm1.2}$ & 55.9$_{\pm0.9}$ & 90.4$_{\pm0.4}$ & 0.5$_{\pm1.3}$ & 71.7$_{\pm1.0}$ & 89.9$_{\pm0.1}$ & 6.4$_{\pm1.8}$ \\
Top-1@0.9    & 39.1$_{\pm0.2}$ & 58.6$_{\pm1.0}$ & 3.3$_{\pm1.1}$ & 62.8$_{\pm0.2}$ & 56.4$_{\pm0.6}$ & 7.1$_{\pm0.6}$ & 55.2$_{\pm0.9}$ & 67.8$_{\pm0.8}$ & $-2.1$$_{\pm0.5}$ & 69.5$_{\pm0.7}$ & 71.6$_{\pm2.0}$ & $-5.6$$_{\pm1.8}$ \\
Entropy@0.3  & 39.0$_{\pm0.1}$ & 51.4$_{\pm1.0}$ & 2.5$_{\pm0.9}$ & 62.9$_{\pm0.2}$ & 46.1$_{\pm0.2}$ & 5.9$_{\pm0.6}$ & 55.3$_{\pm0.9}$ & 62.9$_{\pm1.0}$ & $-1.4$$_{\pm0.5}$ & 69.3$_{\pm0.6}$ & 66.4$_{\pm1.8}$ & $-5.4$$_{\pm0.9}$ \\
Entropy@0.7  & 39.5$_{\pm0.6}$ & 81.5$_{\pm0.6}$ & 9.3$_{\pm1.5}$ & 62.5$_{\pm0.2}$ & 77.4$_{\pm0.5}$ & 12.3$_{\pm1.4}$ & 56.1$_{\pm0.8}$ & 93.0$_{\pm0.0}$ & 2.8$_{\pm1.4}$ & 71.7$_{\pm1.2}$ & 89.5$_{\pm0.7}$ & 6.3$_{\pm6.1}$ \\
\rowcolor{blue!6}
\ours        & 42.6$_{\pm0.6}$ & 67.9$_{\pm1.8}$ & \textbf{15.3$_{\pm1.6}$} & 62.7$_{\pm0.7}$ & 94.5$_{\pm0.2}$ & \textbf{54.6$_{\pm10.0}$} & 57.8$_{\pm1.3}$ & 93.0$_{\pm0.1}$ & \textbf{27.4$_{\pm7.7}$} & 75.2$_{\pm1.3}$ & 84.9$_{\pm0.5}$ & \textbf{27.3$_{\pm3.2}$} \\
\bottomrule
\end{tabular}}
\end{table}

\noindent\textbf{Implementation Details}
\label{sec:implementation}
Experiments were performed using four NVIDIA RTX A6000 GPUs. We framed the emotion recognition task as a multiple-choice question-answering problem, utilizing the probability of the index as the final prediction score. Audio is loaded at 16~kHz mono and passed as raw waveform to the model's audio encoder, and video is sampled at fps=2.0 and resized to $280\times 280$. The $\alpha_1$ and $\alpha_2$ are searched in $\{0.1, 0.3, 0.5, 0.7, 0.9\}$ on the calibration set.

\subsection{Main Results}
\label{ssec:main_results}
Table~\ref{tab:main_results} compares \ours against established baselines across four benchmarks, utilizing three Omni-LLMs. \ours consistently achieves the highest time utility on 11 of 12 model$\times$dataset settings, while attaining wF1 comparable to or exceeding the strongest baseline in most settings. For instance, on the IEMOCAP dataset using Qwen2.5-Omni-7B, \ours improves the weighted-F1 score by 5.3 absolute points (a 13.0\% relative gain) compared to the full-modality baseline. This demonstrates that our method successfully reduces computation time while maintaining, and often enhancing, overall predictive accuracy. These results reveal accuracy gains across most individual emotion categories, confirming that our approach broadly benefits emotion recognition rather than overfitting to a majority class. 

\noindent\textbf{Fine-Grained Performance.}
\label{sec: fine}
To check whether the gains of \ours come from a few easy classes or are spread across the label space, we break down the weighted-F1 by emotion category on IEMOCAP and MELD with Qwen3-Omni-30B, and compare against the text-only baseline in Table~\ref{tab:per_class}. The improvements are consistent rather than concentrated. On IEMOCAP, every one of the six categories improves, with the largest gain on sad and solid gains on neutral, frustrated, and angry, all of which are emotions where vocal prosody carries information that the transcript alone tends to miss. The picture on MELD is similar in shape but smaller in magnitude, which matches the routing distribution in Figure~\ref{fig:routing_analysis}(a): a non-trivial fraction of MELD utterances exit at Tier 1 because the text is already expressive, so there is simply less headroom for the extra modalities to contribute. Even so, every category improves, and the rarer, harder classes such as fear (+9.1) and anger (+8.1) benefit the most. Taken together, these results suggest that \ours is not trading minority-class accuracy for majority-class accuracy; it is recovering predictions across the board, and especially on the categories where text alone is known to be unreliable.

\begin{table}[t]
\centering
\caption{Fine-grained performance comparison between the text-only modality and \ours{} on IEMOCAP and MELD with Qwen3-Omni-30B. Per-class F1 is reported in the per-emotion columns. $\Delta$ denotes the absolute gain.}
\label{tab:per_class}
\small
\begin{subtable}[h]{\linewidth}
\centering
    \caption{IEMOCAP}
    {%
        \centering
       \begin{tabular}{ccccccccc}
\toprule
Methods       & Exc & Fru  & Sad & Neu & Ang & Hap & Avg & W-f1\\ \midrule
T (text-only)  & 35.50 & 43.29 & 40.80 & 49.09 & 37.69 & 36.30 & 40.45 & 41.65 \\
\textbf{\ours} & \textbf{39.33} & \textbf{50.72} & \textbf{57.50} & \textbf{57.17} & \textbf{43.58} & \textbf{41.58} & \textbf{48.31} & \textbf{49.62} \\
$\Delta$       & $+3.83$ & $+7.43$ & $+16.70$ & $+8.08$ & $+5.89$ & $+5.28$ & $+7.87$ & $+7.97$ \\
\bottomrule
\end{tabular}%
}
\end{subtable}

\vspace{6pt}
\begin{subtable}[h]{\linewidth}
\centering
\caption{MELD}
{%
    \centering
   \begin{tabular}{cccccccccc}
\toprule
Methods       & Fear & Neu  & Ang & Sad & Dis & Surp & Joy & Avg & W-f1\\ \midrule
T (text-only)  & 22.99 & 73.37 & 40.23 & 42.78 & 40.00 & 51.11 & 59.69 & 47.17 & 60.24 \\
\textbf{\ours} & \textbf{32.12} & \textbf{74.23} & \textbf{48.31} & \textbf{44.69} & \textbf{41.54} & \textbf{53.44} & \textbf{60.22} & \textbf{50.65} & \textbf{62.42} \\
$\Delta$       & $+9.13$ & $+0.86$ & $+8.08$ & $+1.91$ & $+1.54$ & $+2.33$ & $+0.53$ & $+3.48$ & $+2.18$ \\
\bottomrule
\end{tabular}%
}
\end{subtable}
\end{table}
\begin{figure}[t]
\centering
\includegraphics[width=\textwidth]{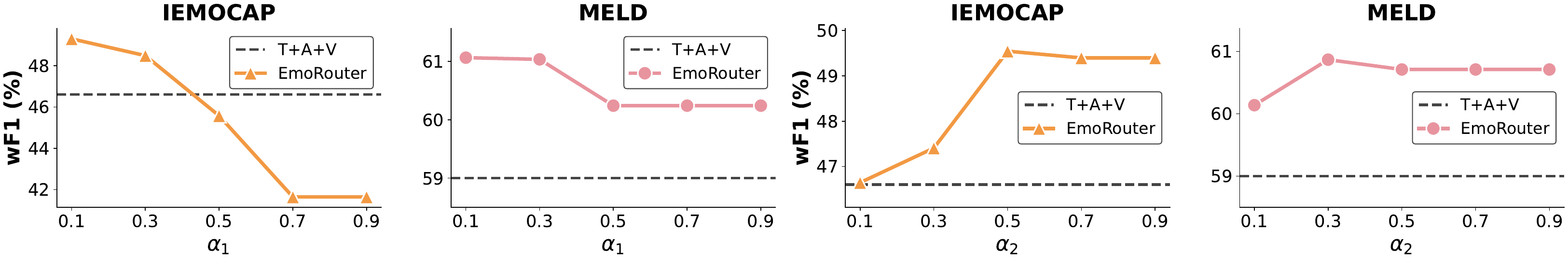}
\caption{wF1 of \ours{} as $\alpha_1$ (left two panels) and $\alpha_2$ (right two panels) vary on IEMOCAP and MELD. When varying $\alpha_1$, $\alpha_2$ is fixed at its CSC-selected value, and vice versa. Dashed grey line marks the T+A+V performance.}
\label{fig:alpha2_sensitivity}
\vspace{-10pt}
\end{figure}
\subsection{Impact of Error Budgets}
\label{ssec:sensitivity}
We analyze the effect of error budget on Qwen3-Omni-30B by manually selecting $\alpha$ without employing CSC, letting $\alpha_1$ and $\alpha_2$ vary independently in the interval $[0.1, 0.9]$. For $\alpha_1$, which is the criterion for exiting the text tier, the weighted-F1 score on MELD remains relatively stable whereas the performance on IEMOCAP drops significantly with the increase of $\alpha_1$, because a permissive text tier allows early exits for utterances where audio information is essential. This demonstrates a high level of variability in the optimal $\alpha_1$ for different benchmarks and explains why a single error threshold does not work. For $\alpha_2$, the maximum performance is achieved at some intermediate value because both low and high $\alpha_2$ lead to undesirable consequences, while the optimal $\alpha_2$ balances between them.

\subsection{Impact of Calibration Set Size}
\label{ssec:calibration_size}
\begin{figure}[t]
\centering
\includegraphics[width=0.65\linewidth]{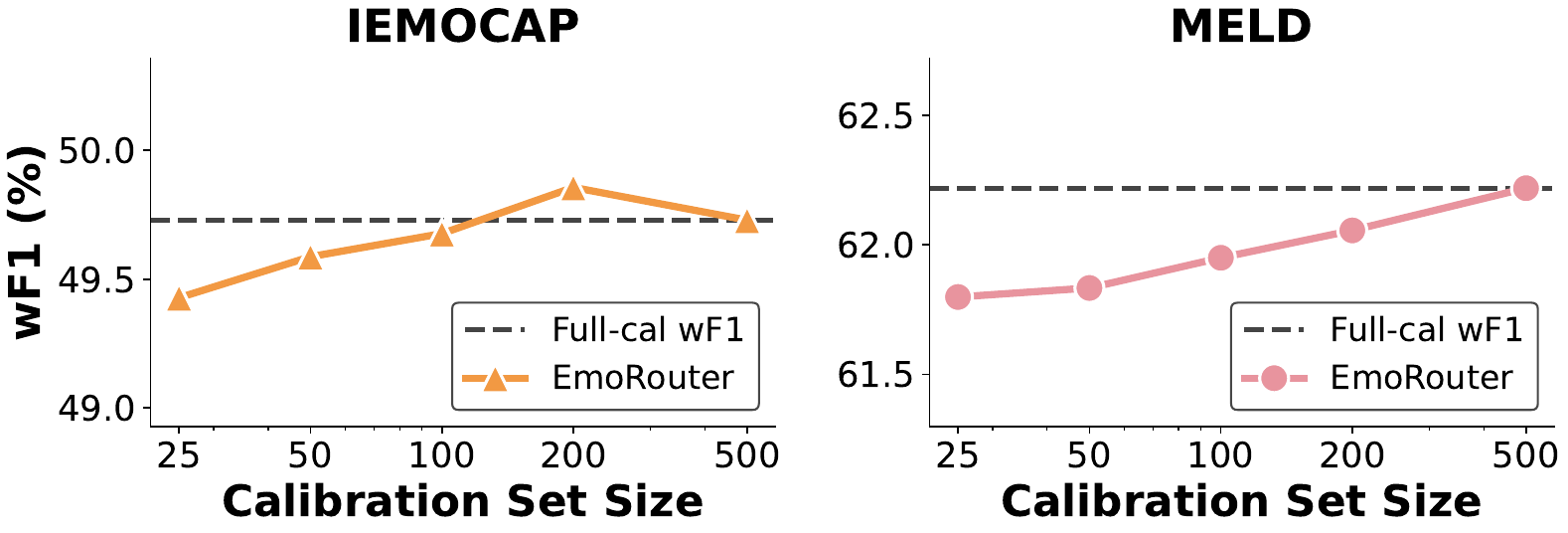}
\caption{wF1 versus calibration set size on IEMOCAP and MELD with Qwen3-Omni-30B.}
\label{fig:cal_size}
\end{figure}

A practical emotion recognition system must balance reliable performance against the annotation cost of calibration data. To investigate the impact of data scarcity on routing stability, we evaluate \ours{} with varying calibration set sizes from 25 to 500 using Qwen3-Omni-30B and compare against the full-calibration reference. The performance curves in Figure~\ref{fig:cal_size} illustrate a rapid convergence of the weighted-F1 score as the available data increases. When the calibration set is extremely small (e.g., $n=25$), \ours{} exhibits mild performance variance, reflecting the statistical instability of quantile estimation at this scale. As the sample size increases to merely 50, \ours{} quickly stabilizes, converging to within $\pm 0.5$ weighted-F1 points of the full-calibration performance. In particular, at a highly constrained annotation budget of 100 samples, \ours performs closely compared to using the full-calibration reference. This indicates that \ours{} achieves robust uncertainty calibration with minimal data requirements, providing practitioners with a highly deployable framework that requires only a few dozen labeled utterances to adapt to a new domain.

\subsection{Modality Routing Analysis}
\label{ssec:routing}
\begin{figure}[t]
\centering
\includegraphics[width=0.9\linewidth]{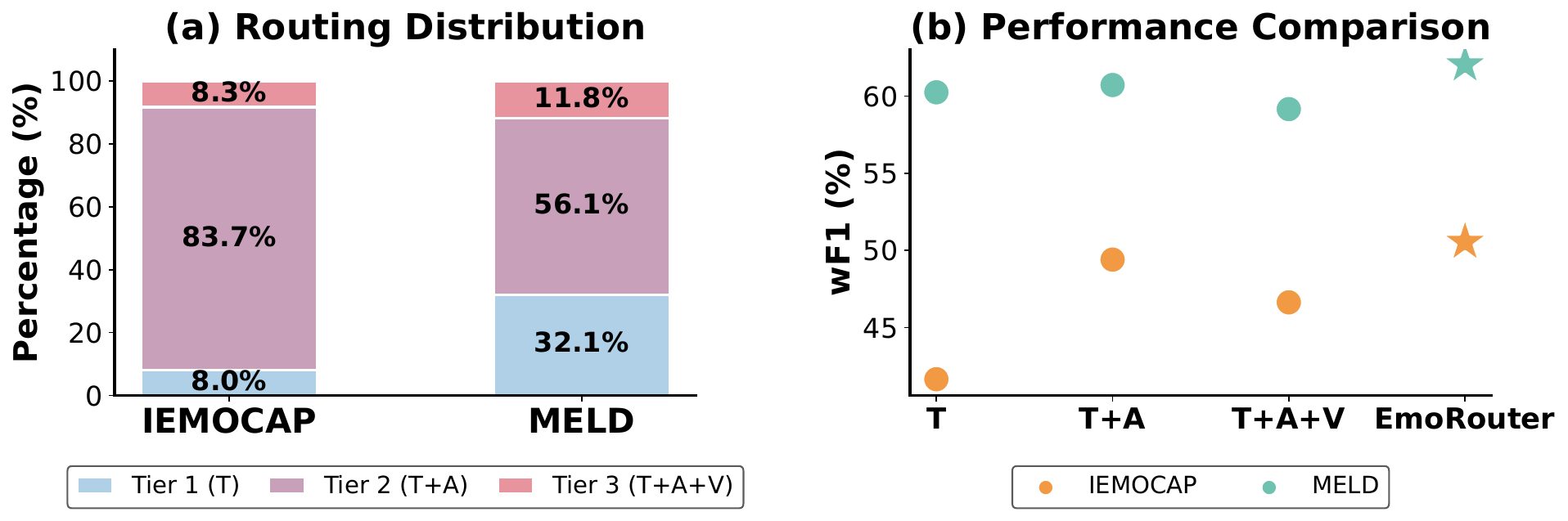}
\caption{Modality Routing Analysis. (a) Modality routing distribution and (b) performance comparison of \ours across different tiers.}
\label{fig:routing_analysis}
\vspace{-10pt}
\end{figure}
To understand how \ours dynamically allocates utterances across different modality tiers, we analyze the proportion of test samples routed to each tier on IEMOCAP and MELD.
As shown in Figure~\ref{fig:routing_analysis} (a), \ours learns adaptive routing strategies that align closely with the distinct characteristics of each dataset. For IEMOCAP, the vast majority of samples are routed to Tier 2 (T+A), with only a small fraction exiting at the text tier or escalating to the full modalities. This indicates that the routing rule frequently defers text-tier predictions to T+A on IEMOCAP. By contrast, MELD exhibits a much higher text-only exit rate (32.1\%), indicating that more samples satisfy the exit criterion using text alone. Meanwhile, 56.1\% of MELD samples exit at Tier 2 (T+A), and 11.8\% proceed to the final T+A+V tier. Crucially, as depicted in Figure~\ref{fig:routing_analysis} (b), this dynamic distribution allows \ours not only to bypass redundant computation but also to outperform any fixed-modality baseline consistently. These results demonstrate that \ours{} adapts its exit decisions to sample-level differences in prediction outcomes relative to the dataset labels, improving both predictive performance and efficiency.

\subsection{Calibration-Free Cross-Model Threshold Transfer}
\label{ssec:cross_model}

To evaluate cross-model transferability, we apply the conformal quantile thresholds $\hat q_\ell$ calibrated on a source model directly to a target model. The source error budgets $\alpha_\ell^*$ are selected by CSC and used to compute these thresholds on the source calibration set. At the target model, prediction sets are constructed using the transferred $\hat q_\ell$, without recomputing quantiles or using target calibration data. Thus, Table~\ref{tab:cross_model} evaluates calibration-free threshold transfer. Transferring only $\alpha_\ell^*$ and recomputing target-side quantiles would constitute a different, error-budget transfer setting.

As shown in Table~\ref{tab:cross_model}, the transferred thresholds generally maintain performance close to native calibration. On MELD, thresholds from Qwen3-Omni-30B improve wF1 on both smaller target models. These results support the empirical transferability of the calibrated routing thresholds across the evaluated backbones without target-side recalibration.

\begin{table}[t]
\centering
\small
\setlength{\tabcolsep}{4pt}
\caption{Results of calibration-free cross-model quantile-threshold transfer on wF1. Rows indicate the source model whose $\hat q_\ell$ thresholds are transferred, and columns indicate the target model. Off-diagonal entries use no target calibration data; diagonal entries use native calibration.}
\label{tab:cross_model}

\begin{minipage}[t]{0.49\linewidth}
\centering
(a) IEMOCAP\\[3pt]
\begin{tabular}{l|ccc}
\toprule
src $\backslash$ tgt & \makecell{Qwen2.5-\\Omni-7B} & \makecell{Qwen3-\\Omni-30B} & \makecell{MiniCPM-\\o-2.6} \\
\midrule
Qwen2.5-Omni-7B & \cellcolor{blue!10}\textbf{45.9} & 49.2 & 41.4 \\
Qwen3-Omni-30B  & 45.1 & \cellcolor{blue!10}\textbf{49.6} & 41.6 \\
MiniCPM-o-2.6   & 45.7 & 49.3 & \cellcolor{blue!10}\textbf{41.8} \\
\bottomrule
\end{tabular}
\end{minipage}%
\hfill
\begin{minipage}[t]{0.49\linewidth}
\centering
(b) MELD\\[3pt]
\begin{tabular}{l|ccc}
\toprule
src $\backslash$ tgt & \makecell{Qwen2.5-\\Omni-7B} & \makecell{Qwen3-\\Omni-30B} & \makecell{MiniCPM-\\o-2.6} \\
\midrule
Qwen2.5-Omni-7B & \cellcolor{blue!10}64.2 & 60.7 & 62.8 \\
Qwen3-Omni-30B  & \textbf{64.4} & \cellcolor{blue!10}\textbf{62.4} & \textbf{63.5} \\
MiniCPM-o-2.6   & 63.3 & 60.7 & \cellcolor{blue!10}62.8 \\
\bottomrule
\end{tabular}
\end{minipage}
\end{table}

\begin{table}[t]
\centering
\caption{Ablation study on Qwen3-Omni-30B, comparing fixed error budgets and alternative tier designs with CSC-tuned \ours{}. The \ours{} row reports the mean results from Table~1. Best $U_t$ per dataset is in \textbf{bold}.}
\label{tab:fixed_alpha_ablation}
\small
\setlength{\tabcolsep}{4pt}
\renewcommand{\arraystretch}{1.05}
\resizebox{0.9\textwidth}{!}{%
\begin{tabular}{l *{4}{rrr}}
\toprule
& \multicolumn{3}{c}{\textbf{IEMOCAP}}
& \multicolumn{3}{c}{\textbf{MELD}}
& \multicolumn{3}{c}{\textbf{MOSEI}}
& \multicolumn{3}{c}{\textbf{MOSI}} \\
\cmidrule(lr){2-4}\cmidrule(lr){5-7}
\cmidrule(lr){8-10}\cmidrule(lr){11-13}
Setting
& wF1 & LRR & $U_t$
& wF1 & LRR & $U_t$
& wF1 & LRR & $U_t$
& wF1 & LRR & $U_t$ \\
\midrule
\rowcolor{gray!12}
\multicolumn{13}{c}{\textit{Ablation on Automatic Error Rate Selection}} \\
$\alpha_1=\alpha_2=0.1$
& 46.3 & 15.7 & 5.5
& 61.1 & 27.7 & 1.2
& 68.7 & 34.3 & 2.7
& 83.5 & 77.4 & \textbf{6.0} \\
$\alpha_1=\alpha_2=0.5$
& 45.6 & 96.4 & 108.4
& 60.3 & 95.0 & 0.0
& 67.1 & 90.0 & 0.0
& 82.1 & 96.3 & 0.0 \\
$\alpha_1=\alpha_2=0.9$
& 41.7 & 98.1 & 0.0
& 60.3 & 95.0 & 0.0
& 67.1 & 90.0 & 0.0
& 82.1 & 96.3 & 0.0 \\
\midrule
\rowcolor{gray!12}
\multicolumn{13}{c}{\textit{Ablation on Modalities and Tier Designs}} \\
w/o T+A
& 46.3 & 10.5 & 5.2
& 60.8 & 22.9 & 0.8
& 68.6 & 30.3 & 2.4
& 83.2 & 69.7 & 3.5 \\
Replace T+A with T+V
& 43.5 & 44.9 & 3.4
& 61.3 & 84.5 & 6.9
& 67.6 & 42.9 & 1.2
& 82.7 & 89.5 & 5.6 \\
\midrule
\rowcolor{blue!6}
\ours{}
& 49.4 & 94.2 & \textbf{142.5}
& 62.7 & 90.4 & \textbf{25.0}
& 68.5 & 84.4 & \textbf{12.7}
& 81.4 & 72.5 & $-0.8$ \\
\bottomrule
\end{tabular}%
}
\end{table}
\subsection{Ablation Study}
We conduct ablations on Qwen3-Omni-30B to examine the contributions of CSC and the modality-tier design. Table~\ref{tab:fixed_alpha_ablation} reports the results.
 
\noindent\textbf{Effect of CSC.} We replace CSC with fixed, shared budgets $\alpha_1 = \alpha_2 \in \{0.1, 0.5, 0.9\}$. A small budget ($\alpha=0.1$) encourages escalation and yields lower LRR than CSC on IEMOCAP, MELD, and MOSEI. Larger budgets enable earlier exits but yield no improvement over text-only inference on MELD, MOSEI, and MOSI ($U_t=0$). CSC achieves the highest time utility on three of the four datasets by selecting per-tier budgets from the calibration set. However, on MOSI, the fixed budget $\alpha=0.1$ achieves higher wF1 and time utility, indicating that maximizing set contraction does not always yield the best accuracy--latency trade-off.
 
\noindent\textbf{Effect of tier design.} Removing T+A leaves only T and T+A+V, requiring samples that do not exit at the text tier to proceed directly to full-modality inference. This reduces LRR on all four benchmarks and lowers wF1 on IEMOCAP and MELD, although wF1 increases slightly on MOSEI and more substantially on MOSI. Replacing T+A with T+V reduces both wF1 and LRR on IEMOCAP, MELD, and MOSEI, but improves both metrics on MOSI. Thus, the audio-intermediate cascade achieves the highest time utility on three datasets, while the MOSI results highlight that the preferred tier design can depend on the dataset.

\section{Conclusion}
We introduce \ours, a training-free framework for cascading adaptive modality in multimodal emotion recognition. Using conformal prediction, \ours{} reliably exits inference at the first confident modality. Our approach outperforms various baselines across 4 widely-used MER benchmarks in both performance and computational efficiency, while generalizing across Omni-LLM backbones without requiring model retraining.

\textbf{Discussion on Broader Impact.} Emotion recognition from speech and video raises privacy concerns,
as these inputs may reveal sensitive personal information.
Deployment should require informed consent and appropriate data
protection. Model predictions may also reflect demographic and
cultural biases in training data, limiting their reliability across
populations and contexts. Since emotional states cannot be reliably
inferred from observable cues alone, \ours{} should not serve as
the sole basis for high-stakes decisions in healthcare, employment,
or law enforcement.

\bibliography{tmlr}

@inproceedings{schuster2021consistent,
  title={Consistent accelerated inference via confident adaptive transformers},
  author={Schuster, Tal and Fisch, Adam and Jaakkola, Tommi and Barzilay, Regina},
  booktitle={Proceedings of the 2021 Conference on Empirical Methods in Natural Language Processing},
  pages={4962--4979},
  year={2021}
}

@article{schuster2022confident,
  title={Confident adaptive language modeling},
  author={Schuster, Tal and Fisch, Adam and Gupta, Jai and Dehghani, Mostafa and Bahri, Dara and Tran, Vinh and Tay, Yi and Metzler, Donald},
  journal={Advances in Neural Information Processing Systems},
  volume={35},
  pages={17456--17472},
  year={2022}
}

@article{hendrycks2016baseline,
  title={A baseline for detecting misclassified and out-of-distribution examples in neural networks},
  author={Hendrycks, Dan and Gimpel, Kevin},
  journal={arXiv preprint arXiv:1610.02136},
  year={2016}
}

@article{gibbs2025conformal,
  title={Conformal prediction with conditional guarantees},
  author={Gibbs, Isaac and Cherian, John J and Cand{\`e}s, Emmanuel J},
  journal={Journal of the Royal Statistical Society Series B: Statistical Methodology},
  volume={87},
  number={4},
  pages={1100--1126},
  year={2025},
  publisher={Oxford University Press UK}
}

@article{fisch2020efficient,
  title={Efficient conformal prediction via cascaded inference with expanded admission},
  author={Fisch, Adam and Schuster, Tal and Jaakkola, Tommi and Barzilay, Regina},
  journal={arXiv preprint arXiv:2007.03114},
  year={2020}
}

@article{barber2023conformal,
  title={Conformal prediction beyond exchangeability},
  author={Barber, Rina Foygel and Candes, Emmanuel J and Ramdas, Aaditya and Tibshirani, Ryan J},
  journal={The Annals of Statistics},
  volume={51},
  number={2},
  pages={816--845},
  year={2023},
  publisher={Institute of Mathematical Statistics}
}

@article{gibbs2021adaptive,
  title={Adaptive conformal inference under distribution shift},
  author={Gibbs, Isaac and Candes, Emmanuel},
  journal={Advances in Neural Information Processing Systems},
  volume={34},
  pages={1660--1672},
  year={2021}
}

@article{jitkrittum2023does,
  title={When does confidence-based cascade deferral suffice?},
  author={Jitkrittum, Wittawat and Gupta, Neha and Menon, Aditya K and Narasimhan, Harikrishna and Rawat, Ankit and Kumar, Sanjiv},
  journal={Advances in Neural Information Processing Systems},
  volume={36},
  pages={9891--9906},
  year={2023}
}

@article{cresswell2024conformal,
  title={Conformal prediction sets improve human decision making},
  author={Cresswell, Jesse C and Sui, Yi and Kumar, Bhargava and Vouitsis, No{\~A}{\c{G}}l},
  journal={arXiv preprint arXiv:2401.13744},
  year={2024}
}

@article{shen2021directed,
  title={Directed acyclic graph network for conversational emotion recognition},
  author={Shen, Weizhou and Wu, Siyue and Yang, Yunyi and Quan, Xiaojun},
  journal={arXiv preprint arXiv:2105.12907},
  year={2021}
}

@article{song2022supervised,
  title={Supervised prototypical contrastive learning for emotion recognition in conversation},
  author={Song, Xiaohui and Huang, Longtao and Xue, Hui and Hu, Songlin},
  journal={arXiv preprint arXiv:2210.08713},
  year={2022}
}

@article{chen2024routerdc,
  title={Routerdc: Query-based router by dual contrastive learning for assembling large language models},
  author={Chen, Shuhao and Jiang, Weisen and Lin, Baijiong and Kwok, James and Zhang, Yu},
  journal={Advances in Neural Information Processing Systems},
  volume={37},
  pages={66305--66328},
  year={2024}
}

@inproceedings{liu2026conformal,
  title={Conformal prediction meets long-tail classification},
  author={Liu, Shuqi and Huang, Jianguo and Ong, Luke},
  booktitle={Proceedings of the AAAI Conference on Artificial Intelligence},
  volume={40},
  number={28},
  pages={23828--23836},
  year={2026}
}

@article{feng2026graphplanner,
  title={GraphPlanner: Graph Memory-Augmented Agentic Routing for Multi-Agent LLMs},
  author={Feng, Tao and Zhang, Haozhen and Lei, Zijie and Han, Peixuan and You, Jiaxuan},
  journal={arXiv preprint arXiv:2604.23626},
  year={2026}
}

@article{liu2026efficient,
  title={Efficient Omni-modal Large Language Model Inference via Per-Query Modality Selection},
  author={Liu, Shuqi and Cao, Yuzhou and Ong, Luke},
  journal={arXiv preprint arXiv:2610.31768},
  year={2026}
}

@inproceedings{su2026cp,
  title={Cp-router: An uncertainty-aware router between llm and lrm},
  author={Su, Jiayuan and Lin, Fulin and Feng, Zhaopeng and Zheng, Han and Wang, Teng and Xiao, Zhenyu and Zhao, Xinlong and Liu, Zuozhu and Cheng, Lu and Wang, Hongwei},
  booktitle={Proceedings of the AAAI Conference on Artificial Intelligence},
  volume={40},
  number={39},
  pages={33065--33073},
  year={2026}
}

@article{hu2022unimse,
  title={Unimse: Towards unified multimodal sentiment analysis and emotion recognition},
  author={Hu, Guimin and Lin, Ting-En and Zhao, Yi and Lu, Guangming and Wu, Yuchuan and Li, Yongbin},
  journal={arXiv preprint arXiv:2211.11256},
  year={2022}
}

@inproceedings{zhang2023dualgats,
  title={Dualgats: Dual graph attention networks for emotion recognition in conversations},
  author={Zhang, Duzhen and Chen, Feilong and Chen, Xiuyi},
  booktitle={Proceedings of the 61st Annual Meeting of the Association for Computational Linguistics (Volume 1: Long Papers)},
  pages={7395--7408},
  year={2023}
}

@article{yu2024emotion,
  title={Emotion-Anchored Contrastive Learning Framework for Emotion Recognition in Conversation},
  author={Yu, Fangxu and Guo, Junjie and Wu, Zhen and Dai, Xinyu},
  journal={arXiv preprint arXiv:2403.20289},
  year={2024}
}

@article{ghosal2019dialoguegcn,
  title={Dialoguegcn: A graph convolutional neural network for emotion recognition in conversation},
  author={Ghosal, Deepanway and Majumder, Navonil and Poria, Soujanya and Chhaya, Niyati and Gelbukh, Alexander},
  journal={arXiv preprint arXiv:1908.11540},
  year={2019}
}

@article{hu2021mmgcn,
  title={Mmgcn: Multimodal fusion via deep graph convolution network for emotion recognition in conversation},
  author={Hu, Jingwen and Liu, Yuchen and Zhao, Jinming and Jin, Qin},
  journal={arXiv preprint arXiv:2107.06779},
  year={2021}
}

@inproceedings{majumder2019dialoguernn,
  title={Dialoguernn: An attentive rnn for emotion detection in conversations},
  author={Majumder, Navonil and Poria, Soujanya and Hazarika, Devamanyu and Mihalcea, Rada and Gelbukh, Alexander and Cambria, Erik},
  booktitle={Proceedings of the AAAI conference on artificial intelligence},
  volume={33},
  number={01},
  pages={6818--6825},
  year={2019}
}

@article{hu2021dialoguecrn,
  title={Dialoguecrn: Contextual reasoning networks for emotion recognition in conversations},
  author={Hu, Dou and Wei, Lingwei and Huai, Xiaoyong},
  journal={arXiv preprint arXiv:2106.01978},
  year={2021}
}

@inproceedings{liu2022dialogueein,
  title={DialogueEIN: Emotion Interaction Network for Dialogue Affective Analysis},
  author={Liu, Yuchen and Zhao, Jinming and Hu, Jingwen and Li, Ruichen and Jin, Qin},
  booktitle={Proceedings of the 29th International Conference on Computational Linguistics},
  pages={684--693},
  year={2022}
}

@article{poria2018meld,
  title={Meld: A multimodal multi-party dataset for emotion recognition in conversations},
  author={Poria, Soujanya and Hazarika, Devamanyu and Majumder, Navonil and Naik, Gautam and Cambria, Erik and Mihalcea, Rada},
  journal={arXiv preprint arXiv:1810.02508},
  year={2018}
}

@inproceedings{ishiwatari2020relation,
  title={Relation-aware graph attention networks with relational position encodings for emotion recognition in conversations},
  author={Ishiwatari, Taichi and Yasuda, Yuki and Miyazaki, Taro and Goto, Jun},
  booktitle={Proceedings of the 2020 Conference on Empirical Methods in Natural Language Processing (EMNLP)},
  pages={7360--7370},
  year={2020}
}

@article{busso2008iemocap,
  title={IEMOCAP: Interactive emotional dyadic motion capture database},
  author={Busso, Carlos and Bulut, Murtaza and Lee, Chi-Chun and Kazemzadeh, Abe and Mower, Emily and Kim, Samuel and Chang, Jeannette N and Lee, Sungbok and Narayanan, Shrikanth S},
  journal={Language resources and evaluation},
  volume={42},
  pages={335--359},
  year={2008},
  publisher={Springer}
}

@article{lewis2019bart,
  title={Bart: Denoising sequence-to-sequence pre-training for natural language generation, translation, and comprehension},
  author={Lewis, Mike and Liu, Yinhan and Goyal, Naman and Ghazvininejad, Marjan and Mohamed, Abdelrahman and Levy, Omer and Stoyanov, Ves and Zettlemoyer, Luke},
  journal={arXiv preprint arXiv:1910.13461},
  year={2019}
}

@article{lei2023instructerc,
  title={Instructerc: Reforming emotion recognition in conversation with a retrieval multi-task llms framework},
  author={Lei, Shanglin and Dong, Guanting and Wang, Xiaoping and Wang, Keheng and Wang, Sirui},
  journal={arXiv preprint arXiv:2309.11911},
  year={2023}
}

@article{zhang2023dialoguellm,
  title={Dialoguellm: Context and emotion knowledge-tuned llama models for emotion recognition in conversations},
  author={Zhang, Yazhou and Wang, Mengyao and Tiwari, Prayag and Li, Qiuchi and Wang, Benyou and Qin, Jing},
  journal={arXiv preprint arXiv:2310.11374},
  year={2023}
}

@article{wu2024beyond,
  title={Beyond Silent Letters: Amplifying LLMs in Emotion Recognition with Vocal Nuances},
  author={Wu, Zehui and Gong, Ziwei and Ai, Lin and Shi, Pengyuan and Donbekci, Kaan and Hirschberg, Julia},
  journal={arXiv preprint arXiv:2407.21315},
  year={2024}
}

@inproceedings{ong2024routellm,
  title={RouteLLM: Learning to Route LLMs from Preference Data},
  author={Ong, Isaac and Almahairi, Amjad and Wu, Vincent and Chiang, Wei-Lin and Wu, Tianhao and Gonzalez, Joseph E and Kadous, M Waleed and Stoica, Ion},
  booktitle={The Thirteenth International Conference on Learning Representations},
year={2024},
}

@article{dey2022conformal,
  title={Conformal prediction for text infilling and part-of-speech prediction},
  author={Dey, Neil and Ding, Jing and Ferrell, Jack and Kapper, Carolina and Lovig, Maxwell and Planchon, Emiliano and Williams, Jonathan P},
  journal={The New England Journal of Statistics in Data Science},
  year={2022}
}

@article{kumar2023conformal,
  title={Conformal prediction with large language models for multi-choice question answering},
  author={Kumar, Bhawesh and Lu, Charlie and Gupta, Gauri and Palepu, Anil and Bellamy, David and Raskar, Ramesh and Beam, Andrew},
  journal={arXiv preprint arXiv:2305.18404},
  year={2023}
}

@article{angelopoulos2021gentle,
  title={A gentle introduction to conformal prediction and distribution-free uncertainty quantification},
  author={Angelopoulos, Anastasios N and Bates, Stephen},
  journal={arXiv preprint arXiv:2107.07511},
  year={2021}
}

@inproceedings{wang2024conu,
  title={ConU: Conformal Uncertainty in Large Language Models with Correctness Coverage Guarantees},
  author={Wang, Zhiyuan and Duan, Jinhao and Cheng, Lu and Zhang, Yue and Wang, Qingni and Shi, Xiaoshuang and Xu, Kaidi and Shen, Heng Tao and Zhu, Xiaofeng},
  booktitle={Findings of the Association for Computational Linguistics: EMNLP 2024},
  pages={6886--6898},
  year={2024}
}

@book{vovk2005algorithmic,
  title={Algorithmic learning in a random world},
  author={Vovk, Vladimir and Gammerman, Alexander and Shafer, Glenn},
  volume={29},
  year={2005},
  publisher={Springer}
}

@inproceedings{maltoudoglou2020bert,
  title={BERT-based conformal predictor for sentiment analysis},
  author={Maltoudoglou, Lysimachos and Paisios, Andreas and Papadopoulos, Harris},
  booktitle={Conformal and Probabilistic Prediction and Applications},
  pages={269--284},
  year={2020},
  organization={PMLR}
}

@InProceedings{pmlr-v204-ernez23a,
  title = 	 {Applying the conformal prediction paradigm for the
 uncertainty quantification of an end-to-end
 automatic speech recognition model (wav2vec 2.0)},
  author =       {Ernez, Fares and Arnold, Alexandre and Galametz, Audrey and Kobus, Catherine and Ould-Amer, Nawal},
  booktitle = 	 {Proceedings of the Twelfth Symposium on Conformal
 and Probabilistic Prediction with Applications},
  pages = 	 {16--35},
  year = 	 {2023},
  editor = 	 {Papadopoulos, Harris and Nguyen, Khuong An and Boström, Henrik and Carlsson, Lars},
  volume = 	 {204},
  series = 	 {Proceedings of Machine Learning Research},
  month = 	 {13--15 Sep},
  publisher =    {PMLR},
  url = 	 {https://proceedings.mlr.press/v204/ernez23a.html}
}

@article{cherian2024large,
  title={Large language model validity via enhanced conformal prediction methods},
  author={Cherian, John and Gibbs, Isaac and Candes, Emmanuel},
  journal={Advances in Neural Information Processing Systems},
  volume={37},
  pages={114812--114842},
  year={2024}
}

@inproceedings{su-etal-2024-api,
    title = "{API} Is Enough: Conformal Prediction for Large Language Models Without Logit-Access",
    author = "Su, Jiayuan  and
      Luo, Jing  and
      Wang, Hongwei  and
      Cheng, Lu",
    editor = "Al-Onaizan, Yaser  and
      Bansal, Mohit  and
      Chen, Yun-Nung",
    booktitle = "Findings of the Association for Computational Linguistics: EMNLP 2024",
    month = nov,
    year = "2024",
    address = "Miami, Florida, USA",
    publisher = "Association for Computational Linguistics",
    url = "https://aclanthology.org/2024.findings-emnlp.54/",
    doi = "10.18653/v1/2024.findings-emnlp.54",
    pages = "979--995"
}

@article{quach2023conformal,
    title={Conformal Language Modeling},
    author={Victor Quach and Adam Fisch and Tal Schuster and Adam Yala and Jae Ho Sohn and Tommi S. Jaakkola and Regina Barzilay},
    booktitle={The Twelfth International Conference on Learning Representations},
    year={2024}
}

@article{xu2025qwen3,
  title={Qwen3-omni technical report},
  author={Xu, Jin and Guo, Zhifang and Hu, Hangrui and Chu, Yunfei and Wang, Xiong and He, Jinzheng and Wang, Yuxuan and Shi, Xian and He, Ting and Zhu, Xinfa and others},
  journal={arXiv preprint arXiv:2509.17765},
  year={2025}
}

@inproceedings{li2023qap,
  title={QAP: A quantum-inspired adaptive-priority-learning model for multimodal emotion recognition},
  author={Li, Ziming and Zhou, Yan and Liu, Yaxin and Zhu, Fuqing and Yang, Chuanpeng and Hu, Songlin},
  booktitle={Findings of the Association for Computational Linguistics: ACL 2023},
  pages={12191--12204},
  year={2023}
}

@article{wang2026emotionthinker,
  title={EmotionThinker: Prosody-Aware Reinforcement Learning for Explainable Speech Emotion Reasoning},
  author={Wang, Dingdong and Liu, Shujie and Zhang, Tianhua and Chen, Youjun and Li, Jinyu and Meng, Helen},
  journal={arXiv preprint arXiv:2601.15668},
  year={2026}
}

@inproceedings{chen2025grpo,
  title={GRPO-Guided Modality Selection Enhanced LoRA-Tuned LLMs for Multimodal Emotion Recognition},
  author={Chen, Yang and Yang, Shuwan and Xiang, Yan and Song, Ran and Huang, Yuxin and Yu, Zhengtao},
  booktitle={Findings of the Association for Computational Linguistics: EMNLP 2025},
  pages={19458--19471},
  year={2025}
}

@inproceedings{gao2026b,
  title={B-GRPO: Unsupervised Speech Emotion Recognition Based on Batched-Group Relative Policy Optimization},
  author={Gao, Yingying and Zhang, Shilei and Yang, Runyan and Cui, Zihao and Feng, Junlan},
  booktitle={ICASSP 2026-2026 IEEE International Conference on Acoustics, Speech and Signal Processing (ICASSP)},
  pages={16422--16426},
  year={2026},
  organization={IEEE}
}

@article{lian2025affectgpt,
  title={Affectgpt-r1: Leveraging reinforcement learning for open-vocabulary multimodal emotion recognition},
  author={Lian, Zheng and Zhang, Fan and Zhang, Yazhou and Tao, Jianhua and Liu, Rui and Chen, Haoyu and Li, Xiaobai and He, Bin},
  journal={arXiv preprint arXiv:2508.01318},
  year={2025}
}

@article{yang2025omni,
  title={Omni-emotion: Extending video mllm with detailed face and audio modeling for multimodal emotion analysis},
  author={Yang, Qize and Bai, Detao and Peng, Yi-Xing and Wei, Xihan},
  journal={arXiv preprint arXiv:2501.09502},
  year={2025}
}

@inproceedings{zheng2023facial,
  title={A facial expression-aware multimodal multi-task learning framework for emotion recognition in multi-party conversations},
  author={Zheng, Wenjie and Yu, Jianfei and Xia, Rui and Wang, Shijin},
  booktitle={Proceedings of the 61st Annual Meeting of the Association for Computational Linguistics (Volume 1: Long Papers)},
  pages={15445--15459},
  year={2023}
}

@article{murzaku2025omnivox,
  title={Omnivox: Zero-shot emotion recognition with omni-llms},
  author={Murzaku, John and Rambow, Owen},
  journal={arXiv preprint arXiv:2503.21480},
  year={2025}
}

@article{zhao2025r1,
  title={R1-omni: Explainable omni-multimodal emotion recognition with reinforcement learning},
  author={Zhao, Jiaxing and Wei, Xihan and Bo, Liefeng},
  journal={arXiv preprint arXiv:2503.05379},
  year={2025}
}

@article{yao2024minicpm,
  title={MiniCPM-V: A GPT-4V Level MLLM on Your Phone},
  author={Yao, Yuan and Yu, Tianyu and Zhang, Ao and Wang, Chongyi and Cui, Junbo and Zhu, Hongji and Cai, Tianchi and Li, Haoyu and Zhao, Weilin and He, Zhihui and others},
  journal={arXiv preprint arXiv:2408.01800},
  year={2024}
}

@inproceedings{zadeh2017tensor,
  title={Tensor fusion network for multimodal sentiment analysis},
  author={Zadeh, Amir and Chen, Minghai and Poria, Soujanya and Cambria, Erik and Morency, Louis-Philippe},
  booktitle={Proceedings of the 2017 conference on empirical methods in natural language processing},
  pages={1103--1114},
  year={2017}
}

@inproceedings{zadeh2018multimodal,
  title={Multimodal language analysis in the wild: Cmu-mosei dataset and interpretable dynamic fusion graph},
  author={Zadeh, AmirAli Bagher and Liang, Paul Pu and Poria, Soujanya and Cambria, Erik and Morency, Louis-Philippe},
  booktitle={Proceedings of the 56th Annual Meeting of the Association for Computational Linguistics (Volume 1: Long Papers)},
  pages={2236--2246},
  year={2018}
}

@inproceedings{yu2025beyond,
  title={Beyond Verbal Cues: Emotional Contagion Graph Network for Causal Emotion Entailment},
  author={Yu, Fangxu and Guo, Junjie and Wu, Zhen and Dai, Xinyu},
  booktitle={Findings of the Association for Computational Linguistics: ACL 2025},
  pages={1755--1767},
  year={2025}
}

@inproceedings{ding2024hybrid,
  title={Hybrid llm: Cost-efficient and quality-aware query routing},
  author={Ding, Dujian and Mallick, Ankur and Wang, Chi and Sim, Robert and Mukherjee, Subhabrata and R{\"u}hle, Victor and Lakshmanan, Laks and Awadallah, Ahmed H},
  booktitle={International Conference on Learning Representations},
  volume={2024},
  pages={41348--41366},
  year={2024}
}

@article{feng2026llmrouter,
  title={LLMRouter: Unified infrastructure for developing, evaluating, and deploying LLM routers},
  author={Feng, Tao and Yu, Fangxu and Zhang, Haozhen and Dai, Zhongjie and Yuan, Liangqi and Lei, Zijie and Zhang, Weizhi and Zhu, Kunlun and Yue, Haodong and Xuan, Keyang and others},
  journal={arXiv preprint arXiv:2608.06867},
  year={2026}
}

@article{yu2026tsrouter,
  title={TSRouter: Dynamic Modality-Model Selection for Time Series Reasoning},
  author={Yu, Fangxu and Feng, Tao and Min, Dehai and Cheng, Lu and Liu, Ge and Zhou, Tianyi},
  journal={arXiv preprint arXiv:2607.08940},
  year={2026}
}

@inproceedings{silva2025conformal,
  title={Conformal prediction for zero-shot models},
  author={Silva-Rodr{\'\i}guez, Julio and Ben Ayed, Ismail and Dolz, Jose},
  booktitle={Proceedings of the IEEE/CVF Conference on Computer Vision and Pattern Recognition},
  pages={19931--19941},
  year={2025}
}

@inproceedings{radford2023robust,
  title={Robust speech recognition via large-scale weak supervision},
  author={Radford, Alec and Kim, Jong Wook and Xu, Tao and Brockman, Greg and McLeavey, Christine and Sutskever, Ilya},
  booktitle={International conference on machine learning},
  pages={28492--28518},
  year={2023},
  organization={PMLR}
}

@article{gauthier2026backward,
  title={Backward conformal prediction},
  author={Gauthier, Etienne and Bach, Francis and Jordan, Michael},
  journal={Advances in Neural Information Processing Systems},
  volume={38},
  pages={103897--103929},
  year={2026}
}

@article{angelopoulos2024theoretical,
  title={Theoretical foundations of conformal prediction},
  author={Angelopoulos, Anastasios N and Barber, Rina Foygel and Bates, Stephen},
  journal={arXiv preprint arXiv:2411.11824},
  year={2024}
}
\bibliographystyle{tmlr}

\appendix
\newpage

\section{Pseudocode}
\label{sec: pseudocode}
We provide the pseudocode of \ours in Algorithm~\ref{alg:emorouter}.

\begin{algorithm}[t]
\small
\caption{\ours{}: CSC calibration and cascade routing}
\label{alg:emorouter}
\begin{algorithmic}[1]
\Require Calibration set $\mathcal{D}_{\mathrm{cal}}=\{(x_i,y_i)\}_{i=1}^{n}$; Omni-LLM $\mathcal{M}$ with $L$ modality tiers; grid $\mathcal{A}$
\Statex \textbf{Stage 1: Offline calibration via CSC}
\For{each tier $\ell \in \{1,\dots,L\}$ and each $x_i \in \mathcal{D}_{\mathrm{cal}}$}
    \State Compute $\hat{\pi}^{(\ell)}(x_i) \gets \mathcal{M}(x_i; \mathcal{T}_\ell)$
\EndFor
\For{each candidate $(\alpha_1,\dots,\alpha_{L-1}) \in \mathcal{A}$}
    \For{each non-terminal tier $\ell$}
        \State $s_i^{(\ell)} \gets 1 - \hat{\pi}^{(\ell)}_{y_i}(x_i)$ for all $i$
        \State $\hat{q}_\ell \gets \lceil(n{+}1)(1{-}\alpha_\ell)\rceil/n$ quantile of $\{s_i^{(\ell)}\}$
        \State $\mathcal{C}_{\alpha_\ell}^{(\ell)}(x_i) \gets \{y : \hat{\pi}_y^{(\ell)}(x_i) \geq 1{-}\hat{q}_\ell\}$
    \EndFor
    \State Compute $\mathrm{CSC}(\alpha_1,\dots,\alpha_{L-1})$ via~\eqref{eq:CSC}
\EndFor
\State $(\alpha_1^*,\dots,\alpha_{L-1}^*) \gets \argmax_{\alpha\in\mathcal{A}} \mathrm{CSC}(\alpha)$
\Statex \textbf{Stage 2: Test-time cascade routing for input $x$}
\For{$\ell = 1$ to $L-1$}
    \State Run $\mathcal{M}$ at tier $\mathcal{T}_\ell$ to obtain $\hat{\pi}^{(\ell)}(x)$
    \State Build $\mathcal{C}_{\alpha_\ell^*}^{(\ell)}(x)$ using $\hat{q}_\ell$ from Stage 1
    \If{$|\mathcal{C}_{\alpha_\ell^*}^{(\ell)}(x)| \leq 1$}
        \State \Return $\arg\max_y \hat{\pi}_y^{(\ell)}(x)$
    \EndIf
\EndFor
\State Run $\mathcal{M}$ at the final tier $\mathcal{T}_L$
\State \Return $\arg\max_y \hat{\pi}_y^{(L)}(x)$
\end{algorithmic}
\end{algorithm}

\section{Dataset Details}
\label{sec:dataset_details}
We provide detailed descriptions of the 4 benchmarks below and summarize their statistics in Table~\ref{tab:dataset_stats}. The benchmarks cover complementary prediction tasks: the 2 conversational datasets (IEMOCAP and MELD) require distinguishing fine-grained emotion categories in context-dependent utterances, whereas the 2 opinion datasets (MOSEI and MOSI) require judging sentiment polarity in single-speaker monologues, where the transcript often carries explicit evaluative cues.

\begin{table}[t]
\centering
\small
\caption{Overview of the four MER benchmarks.}
\label{tab:dataset_stats}
\begin{tabular}{lllcc}
\toprule
Dataset & Source & Interaction & Label Space & \#Classes \\
\midrule
IEMOCAP & Acted sessions & Dyadic & Emotion & 6 \\
MELD & TV series (\textit{Friends}) & Multi-party & Emotion & 7 \\
MOSEI & YouTube videos & Monologue & Sentiment & 2 \\
MOSI & YouTube movie reviews & Monologue & Sentiment & 2 \\
\bottomrule
\end{tabular}
\end{table}

\noindent\textbf{IEMOCAP.} IEMOCAP~\citep{busso2008iemocap} contains approximately 12 hours of audio-visual recordings of dyadic conversations performed by 10 professional actors (5 female and 5 male) across 5 sessions, each involving one pair of speakers in scripted and improvised scenarios designed to elicit emotional expressions. Following standard practice~\citep{majumder2019dialoguernn, hu2021mmgcn}, we use 6 emotion categories: Happy, Sad, Neutral, Angry, Excited, and Frustrated, and adopt the conventional speaker-independent split~\citep{chen2025grpo}, in which Sessions 1--4 provide 5,163 training and 647 validation utterances and Session 5 provides 1,623 test utterances. Several category pairs (e.g., Happy vs.\ Excited and Angry vs.\ Frustrated) share similar lexical content but differ in vocal arousal, which makes IEMOCAP particularly challenging for text-only prediction. This is consistent with the routing analysis in Figure~\ref{fig:routing_analysis}, where the vast majority of IEMOCAP utterances exit at Tier 2 (T+A).

\noindent\textbf{MELD.} MELD~\citep{poria2018meld} extends the text-only EmotionLines dataset with audio and visual modalities and contains 13,708 utterances from 1,433 multi-party dialogues in the TV series \textit{Friends}. Each utterance is labeled with one of 7 emotions: Anger, Disgust, Fear, Joy, Neutral, Sadness, and Surprise. The label distribution is highly imbalanced, with Neutral accounting for nearly half of all utterances while Fear and Disgust each account for less than 3\%, which is why we report weighted-F1 and additionally break down performance by category in Table~\ref{tab:per_class}. We use the official split of 9,989, 1,109, and 2,610 utterances for training, validation, and testing, respectively.

\noindent\textbf{CMU-MOSEI.} CMU-MOSEI~\citep{zadeh2018multimodal} is one of the largest sentence-level multimodal sentiment datasets, containing 23,453 annotated video segments from more than 1,000 speakers across 250 topics on YouTube. Each segment is annotated with a sentiment intensity score ranging from $-3$ (highly negative) to $+3$ (highly positive), which we binarize into Positive ($>0$) and Negative ($<0$) following MER-UniBench~\citep{lian2025affectgpt}. The broad coverage of speakers, topics, and recording conditions makes MOSEI a stress test for whether \ours generalizes beyond acted conversations. We use the standard split of 16,326, 1,871, and 4,659 segments for training, validation, and testing, respectively.

\noindent\textbf{CMU-MOSI.} CMU-MOSI~\citep{zadeh2017tensor} contains 2,199 opinion segments from 93 YouTube movie review videos by 89 speakers, each annotated with a sentiment intensity score in $[-3, 3]$ and binarized in the same way as MOSEI. Its small scale and single-domain focus complement MOSEI. Consistent with the prevalence of explicit opinion words in movie reviews, adding audio and video to the transcript even lowers wF1 on MOSI with both Qwen backbones (Table~\ref{tab:main_results}). We use the standard split of 1,284, 229, and 686 segments for training, validation, and testing, respectively.

\noindent\textbf{Data Usage.} Since \ours is training-free, no split is used to update model parameters. We draw the calibration set $\mathcal{D}_{\mathrm{cal}}$ from the validation split of each dataset, which serves as the full-calibration reference in \S\ref{ssec:calibration_size}, and report all results on the test split.

\section{Baseline Design and the Contribution of CSC}
\label{sec:baseline_discussion}

All routing baselines in Table~\ref{tab:main_results} share the identical three-tier cascade of \S\ref{sec:framework} and differ only in the exit signal: Random Routing uses none, SCC uses cross-tier agreement, and Top-1 Probability and Entropy Routing use point confidence. For the latter two, no principled label-free rule exists for choosing the threshold, which is precisely the problem CSC solves, so we sweep thresholds over the same grid used for $\alpha$ (Section~\ref{sec:implementation}) and report a permissive and a conservative operating point per method (e.g., Top-1@0.7/@0.9) that span the grid's two ends; a confidence baseline whose threshold is instead tuned on the calibration set would select a point from this same grid, and \ours{} outperforms both reported operating points on time utility in 11 of 12 settings. Because the cascade structure is held fixed across all methods, the margins in Table~\ref{tab:main_results} measure the value of the CP set-size signal itself. The fixed-$\alpha$ ablation in Table~\ref{tab:fixed_alpha_ablation} goes one step further by holding even the CP machinery fixed and varying only the budget-selection rule: fixed budgets either over-cascade ($\alpha{=}0.1$) or collapse to text-only routing on three of four datasets ($\alpha \in \{0.5, 0.9\}$), whereas CSC dominates on every dataset. This attributes the remaining gain to per-tier budget selection rather than to the generic cascade, and the sensitivity analysis in \S\ref{ssec:sensitivity} traces how costly a mis-set $\alpha$ is and how strongly its optimum varies across datasets.

Beyond these baselines, two alternative comparators are informative. An \emph{oracle router} that assigns each utterance the cheapest tier whose prediction is correct upper-bounds any per-utterance modality selector, but it requires label access at every tier and is therefore a diagnostic rather than a deployable router. The pilot analysis in Figure~\ref{fig:motivation}~(b) exposes the information such an oracle would exploit, namely the distribution of utterances over the modality combinations that predict them correctly: a large portion of utterances are resolvable at cheap tiers, which is exactly the headroom that \ours{} converts into latency savings. We leave charting the explicit oracle frontier to future work. A \emph{cost-aware trained router}, i.e., a lightweight model mapping utterance features to a tier, is likewise complementary rather than directly comparable: it introduces a supervised training stage and a dataset-specific component, forfeiting the training-free, model-agnostic properties that allow \ours{} to transfer across backbones without recalibration (\S\ref{ssec:cross_model}); the closest trained selector~\citep{chen2025grpo} is discussed in Related Work.

\section{Additional Experimental Results}
\subsection{Performance with ASR}
\label{sec:asr}

Our main experiments use dataset-provided transcripts. To evaluate
\ours{} when transcripts are not readily available, we replace them
with transcripts generated by Whisper-turbo~\citep{radford2023robust} and re-evaluate all tiers
on Qwen2.5-Omni-7B, including ASR time in the latency measurements.
We also compare the default text-first cascade
(T $\rightarrow$ T+A $\rightarrow$ T+A+V) with an audio-first
alternative (A $\rightarrow$ A+T $\rightarrow$ A+T+V).

\begin{table}[t]
\centering
\caption{Results with Whisper-turbo transcripts on Qwen2.5-Omni-7B.
Latency measurements include ASR time. Both cascade orders use the
same text-only reference for $U_t$ and the same full-modality reference
for LRR. Best $U_t$ per dataset is in \textbf{bold}.}
\label{tab:asr_results}
\small
\setlength{\tabcolsep}{4pt}
\renewcommand{\arraystretch}{1.05}
\resizebox{0.88\textwidth}{!}{%
\begin{tabular}{l *{4}{rrr}}
\toprule
& \multicolumn{3}{c}{\textbf{IEMOCAP}}
& \multicolumn{3}{c}{\textbf{MELD}}
& \multicolumn{3}{c}{\textbf{MOSEI}}
& \multicolumn{3}{c}{\textbf{MOSI}} \\
\cmidrule(lr){2-4}
\cmidrule(lr){5-7}
\cmidrule(lr){8-10}
\cmidrule(lr){11-13}
Method
& wF1 & LRR & $U_t$
& wF1 & LRR & $U_t$
& wF1 & LRR & $U_t$
& wF1 & LRR & $U_t$ \\
\midrule
\rowcolor{gray!12}
\multicolumn{13}{c}{\textit{Text-first with ASR}} \\
T
& 40.4 & -- & --
& 55.1 & -- & --
& 58.0 & -- & --
& 81.4 & -- & -- \\
T+A+V
& 41.3 & 0.0 & 1.0
& 58.3 & 0.0 & 3.3
& 59.5 & 0.0 & 1.5
& 81.0 & 0.0 & $-0.4$ \\
\rowcolor{blue!6}
\ours
& 44.8 & 87.5 & \textbf{35.6}
& 59.2 & 92.3 & \textbf{53.4}
& 59.6 & 92.0 & \textbf{20.2}
& 82.5 & 95.5 & \textbf{23.1} \\
\addlinespace[2pt]
\midrule
\rowcolor{gray!12}
\multicolumn{13}{c}{\textit{Audio-first with ASR}} \\
A
& 39.6 & -- & --
& 52.1 & -- & --
& 54.0 & -- & --
& 81.3 & -- & -- \\
T+A+V
& 41.3 & 0.0 & 1.0
& 58.3 & 0.0 & 3.3
& 59.5 & 0.0 & 1.5
& 81.0 & 0.0 & $-0.4$ \\
\rowcolor{blue!6}
\ours
& 45.1 & 84.5 & 30.5
& 59.0 & 89.3 & 36.5
& 59.7 & 86.5 & 12.2
& 82.6 & 91.9 & 14.6 \\
\bottomrule
\end{tabular}%
}
\end{table}

As shown in Table~\ref{tab:asr_results}, text-first \ours{}
outperforms full-modality inference on all four benchmarks while
reducing latency by 87.5--95.5\%, demonstrating that its benefits
persist with automatically generated transcripts.
The two cascade orders achieve comparable wF1, differing by at most
0.3 points, whereas text-first routing consistently achieves higher
LRR and $U_t$. Moreover, text-only inference with ASR transcripts
outperforms audio-only inference on all four benchmarks.
These results support text as a cost-effective first tier even
when transcription cost and potential ASR errors are considered.

\subsection{Prediction-Set Contraction under Shared Error Budgets}
\label{sec:shared_alpha}

CSC measures prediction-set contraction across tiers with potentially
different error budgets. To examine whether contraction also occurs
without changing the budget, we compare the average prediction-set
sizes at the T and T+A tiers under shared budgets
$\alpha \in \{0.1, 0.2, 0.3\}$ on the calibration set using
Qwen2.5-Omni-7B.

\begin{table}[t]
\centering
\small
\setlength{\tabcolsep}{8pt}
\renewcommand{\arraystretch}{1.1}
\caption{Average prediction-set sizes at the T and T+A tiers
under shared error budgets on Qwen2.5-Omni-7B.
Each entry reports the transition from T to T+A.}
\label{tab:shared_alpha}
\begin{tabular}{lccc}
\toprule
\textbf{Dataset}
& $\alpha=0.1$
& $\alpha=0.2$
& $\alpha=0.3$ \\
\midrule
IEMOCAP
& $4.34 \rightarrow 3.23$
& $3.22 \rightarrow 2.22$
& $2.36 \rightarrow 1.71$ \\
MELD
& $2.98 \rightarrow 2.73$
& $2.00 \rightarrow 1.79$
& $1.48 \rightarrow 1.31$ \\
MOSEI
& $2.00 \rightarrow 1.92$
& $1.62 \rightarrow 1.55$
& $1.32 \rightarrow 1.28$ \\
MOSI
& $1.44 \rightarrow 1.30$
& $1.05 \rightarrow 1.01$
& $0.84 \rightarrow 0.80$ \\
\bottomrule
\end{tabular}
\end{table}

Table~\ref{tab:shared_alpha} shows that adding audio reduces the
average prediction-set size on all four benchmarks under every
shared budget. This indicates that the observed contraction is
not solely a consequence of changing $\alpha$, and supports
interpreting it as reduced predictive ambiguity after incorporating
audio at a fixed nominal error level.
Meanwhile, increasing $\alpha$ also shrinks prediction sets within
a fixed tier: on IEMOCAP, the average text-tier set size decreases
from 4.34 to 2.36 as $\alpha$ increases from 0.1 to 0.3.
Consequently, when $\alpha_1 \neq \alpha_2$, CSC reflects the
combined effects of modality augmentation and budget allocation,
rather than an isolated measure of modality-induced uncertainty
reduction.

\subsection{Comparison with Calibration-Tuned Confidence Routing}
\label{sec:tuned_confidence}

To examine whether the gains persist when confidence-based baselines
also receive calibration-based threshold selection, we tune Top-1
Probability and Entropy Routing on the same calibration set as
\ours{}. For each baseline, we select two independent per-tier
thresholds from $\{0.1, 0.3, 0.5, 0.7, 0.9\}$ by maximizing the
average reduction in its corresponding uncertainty score across
consecutive tiers over samples that have not yet exited.
This follows the contraction-based selection protocol of CSC.

\begin{table}[t]
\centering
\caption{Comparison with calibration-tuned confidence routing on
Qwen2.5-Omni-7B. All methods use the same calibration set.
Best $U_t$ per dataset is in \textbf{bold}.}
\label{tab:tuned_confidence}
\small
\setlength{\tabcolsep}{4pt}
\renewcommand{\arraystretch}{1.05}
\resizebox{0.88\textwidth}{!}{%
\begin{tabular}{l *{4}{rrr}}
\toprule
& \multicolumn{3}{c}{\textbf{IEMOCAP}}
& \multicolumn{3}{c}{\textbf{MELD}}
& \multicolumn{3}{c}{\textbf{MOSEI}}
& \multicolumn{3}{c}{\textbf{MOSI}} \\
\cmidrule(lr){2-4}
\cmidrule(lr){5-7}
\cmidrule(lr){8-10}
\cmidrule(lr){11-13}
Method
& wF1 & LRR & $U_t$
& wF1 & LRR & $U_t$
& wF1 & LRR & $U_t$
& wF1 & LRR & $U_t$ \\
\midrule
Top-1 Probability 
& 40.9 & 97.3 & $-0.2$
& 61.9 & 85.5 & 9.0
& 59.5 & 95.9 & 15.5
& 81.4 & 95.5 & 6.7 \\
Entropy 
& 41.7 & 77.0 & 3.5
& 62.5 & 70.3 & 6.3
& 58.8 & 94.6 & $-1.8$
& 81.1 & 91.2 & 0.1 \\
\rowcolor{blue!6}
\ours
& 45.9 & 83.6 & \textbf{30.5}
& 64.2 & 87.2 & \textbf{28.1}
& 61.1 & 92.1 & \textbf{27.9}
& 81.9 & 90.4 & \textbf{8.3} \\
\bottomrule
\end{tabular}%
}
\end{table}

\begin{figure}[t]
\centering
\includegraphics[width=1.0\linewidth]{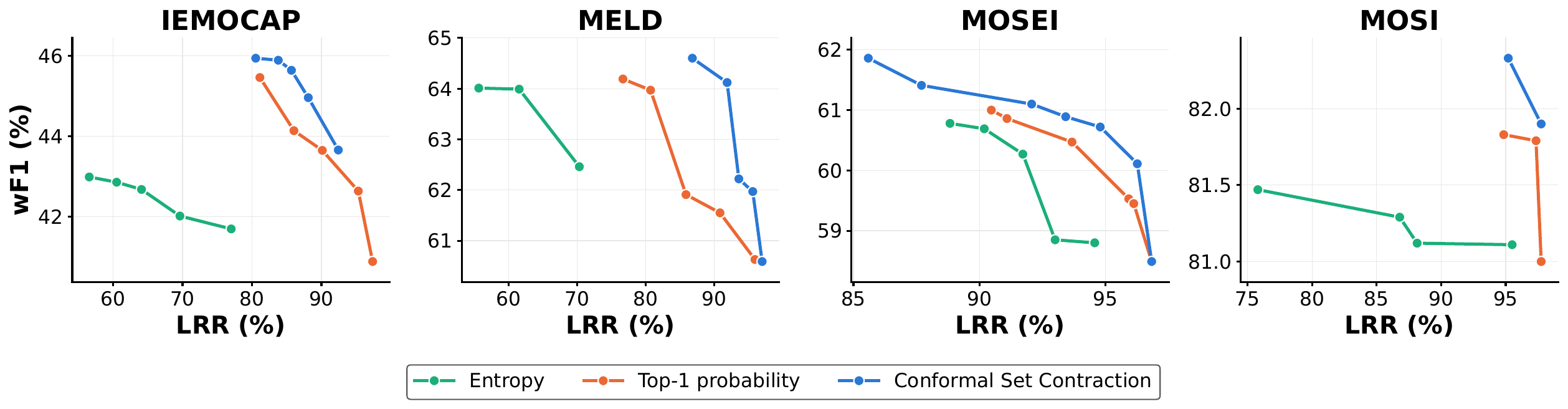}
\caption{wF1--LRR Pareto frontiers on four datasets with Qwen2.5-Omni-7B. All methods are evaluated using the same per-tier threshold grid. Higher wF1 and LRR indicate better predictive performance and greater latency reduction, respectively.}
\label{fig:pareto_frontiers}
\vspace{-10pt}
\end{figure}

As shown in Table~\ref{tab:tuned_confidence}, \ours{} achieves
the highest wF1 and time utility on all four benchmarks. These results support the
effectiveness of conformal prediction-set size combined with CSC
under the matched calibration protocol. We further compare the wF1--LRR Pareto frontiers using the same
per-tier threshold grid for all methods. Figure~\ref{fig:pareto_frontiers} shows that conformal-set-based (CSC)
routing generally achieves higher wF1 at comparable LRR across all four datasets. These results support its effectiveness across a range of accuracy--latency trade-offs.

\subsection{All-Tier Confidence Selection}
\label{sec:maxconf}

To compare sequential routing with confidence selection across all modality tiers, we evaluate MaxConf on Qwen2.5-Omni-7B. For every utterance, MaxConf runs T, T+A, and T+A+V and selects the prediction with the highest calibrated confidence. Its latency includes the cost of all three tiers.

\begin{table}[t]
\centering
\caption{Comparison with all-tier confidence selection on Qwen2.5-Omni-7B. MaxConf evaluates all three tiers for every utterance. Best $U_t$ per dataset is in \textbf{bold}.}
\label{tab:maxconf}
\small
\setlength{\tabcolsep}{4pt}
\renewcommand{\arraystretch}{1.05}
\resizebox{0.88\textwidth}{!}{%
\begin{tabular}{l *{4}{rrr}}
\toprule
& \multicolumn{3}{c}{\textbf{IEMOCAP}}
& \multicolumn{3}{c}{\textbf{MELD}}
& \multicolumn{3}{c}{\textbf{MOSEI}}
& \multicolumn{3}{c}{\textbf{MOSI}} \\
\cmidrule(lr){2-4}\cmidrule(lr){5-7}\cmidrule(lr){8-10}\cmidrule(lr){11-13}
Method & wF1 & LRR & $U_t$ & wF1 & LRR & $U_t$ & wF1 & LRR & $U_t$ & wF1 & LRR & $U_t$ \\
\midrule
MaxConf (all tiers)
& 44.5 & $-11.8$ & 3.2 & 63.9 & $-12.9$ & 2.9
& 61.0 & $-8.3$ & 1.9 & 81.3 & $-9.7$ & 0.2 \\
\rowcolor{blue!6}
\ours
& 45.9 & 83.6 & \textbf{30.5} & 64.2 & 87.2 & \textbf{28.1}
& 61.1 & 92.1 & \textbf{27.9} & 81.9 & 90.4 & \textbf{8.3} \\
\bottomrule
\end{tabular}%
}
\end{table}

Table~\ref{tab:maxconf} shows that \ours{} achieves higher wF1 on all four benchmarks while substantially reducing latency. MaxConf incurs negative LRR because evaluating every tier costs more than running T+A+V alone. These results show that selecting the most confident prediction across all tiers does not recover the accuracy--efficiency trade-off achieved by \ours{}.

\subsection{Comparison with a Cost-Aware Trained Router}
\label{sec:trained_router}

We compare \ours{} with Hybrid-LLM~\citep{ding2024hybrid, feng2026llmrouter}, a representative cost-aware
trained router, by adapting its selection between models to
selection between modality tiers. We train two routers on the
same calibration set used by \ours{}. The trained routing
strategy selects a modality tier directly from the transcript
before running the Omni-LLM. We evaluate both methods on
Qwen2.5-Omni-7B.
As shown in Table~\ref{tab:trained_router}, \ours{} achieves
higher latency reduction and time utility on all four benchmarks,
and higher wF1 on three of them. 
These results demonstrate that \ours{} provides a competitive
accuracy--efficiency trade-off without requiring router training.
\begin{table}[t]
\centering
\caption{Comparison with a cost-aware trained router on
Qwen2.5-Omni-7B. Both methods use the same calibration set.
Best results for each metric are in \textbf{bold}.}
\label{tab:trained_router}
\small
\setlength{\tabcolsep}{4pt}
\renewcommand{\arraystretch}{1.05}
\resizebox{0.88\textwidth}{!}{%
\begin{tabular}{l *{4}{rrr}}
\toprule
& \multicolumn{3}{c}{\textbf{IEMOCAP}}
& \multicolumn{3}{c}{\textbf{MELD}}
& \multicolumn{3}{c}{\textbf{MOSEI}}
& \multicolumn{3}{c}{\textbf{MOSI}} \\
\cmidrule(lr){2-4}
\cmidrule(lr){5-7}
\cmidrule(lr){8-10}
\cmidrule(lr){11-13}
Method
& wF1 & LRR & $U_t$
& wF1 & LRR & $U_t$
& wF1 & LRR & $U_t$ 
& wF1 & LRR & $U_t$ \\
\midrule
Hybrid LLM
& 44.6 & 71.5 & 13.1
& 61.5 & 77.6 & 4.2
& 60.1 & 73.2 & 4.6
& \textbf{82.1} & 70.4 & 3.3 \\
\rowcolor{blue!6}
\ours
& \textbf{45.9} & \textbf{83.6} & \textbf{30.5}
& \textbf{64.2} & \textbf{87.2} & \textbf{28.1}
& \textbf{61.1} & \textbf{92.1} & \textbf{27.9}
& 81.9 & \textbf{90.4} & \textbf{8.3} \\
\bottomrule
\end{tabular}%
}
\end{table}

\subsection{Analysis of Routing Outcomes}
\label{sec:routing_outcomes}

To examine how early exiting and modality escalation contribute
to prediction outcomes, we report four routing diagnostics on
Qwen2.5-Omni-7B:
(i) \emph{correct early exits}, where an utterance exits at
Tier 1 or Tier 2 with a correct prediction;
(ii) \emph{later-tier corrections}, where escalation to a later
tier corrects an earlier incorrect prediction;
(iii) \emph{premature incorrect exits}, where an utterance exits
early with an incorrect prediction that T+A+V would correct;
and (iv) \emph{modality interference}, where escalation changes
an earlier correct prediction into an incorrect one.
Correctness in all four diagnostics is measured against the dataset labels.
All percentages are computed over the full test set.

\begin{table}[t]
\centering
\small
\setlength{\tabcolsep}{6pt}
\renewcommand{\arraystretch}{1.1}
\caption{Routing diagnostics for \ours{} on Qwen2.5-Omni-7B.
Correctness is measured against the dataset labels.
All values are percentages of test utterances. The diagnostics
may overlap and do not constitute an exhaustive partition.}
\label{tab:routing_outcomes}
\begin{tabular}{lcccc}
\toprule
\textbf{Dataset}
& \shortstack{\textbf{Correct}\\\textbf{early exits}}
& \shortstack{\textbf{Later-tier}\\\textbf{corrections}}
& \shortstack{\textbf{Premature}\\\textbf{incorrect exits}}
& \shortstack{\textbf{Modality}\\\textbf{interference}} \\
\midrule
IEMOCAP & 45.7\% & 10.5\% & 4.1\% & 5.8\% \\
MELD    & 63.1\% & 10.3\% & 4.6\% & 6.8\% \\
MOSEI   & 59.3\% &  5.1\% & 2.2\% & 3.2\% \\
MOSI    & 78.3\% &  5.7\% & 2.0\% & 5.3\% \\
\bottomrule
\end{tabular}
\end{table}

Table~\ref{tab:routing_outcomes} shows that a substantial
proportion of test utterances are correctly resolved before
reaching the full-modality tier. Later-tier corrections also
outnumber cases of modality interference on every benchmark,
indicating that escalation recovers the annotated labels for
some utterances that earlier tiers fail to classify correctly.
However, additional modalities can still overturn correct
predictions, and premature exits miss corrections available
at the full-modality tier. Together, these results support
the benefit of adapting the exit tier to individual utterances.

\section{Early-Exit Reliability and Failure Analysis}
\label{sec:failure_cases}
\subsection{Early Exits versus Full-Modality Inference}
To assess the benefits and risks of early exiting, we compare
\ours{} with full-modality inference on Qwen2.5-Omni-7B.
We consider two types of disagreement: (i) \emph{missed corrections},
where \ours{} exits at Tier 1 or Tier 2 with an incorrect prediction
while T+A+V would be correct; and (ii) \emph{avoided errors},
where \ours{} exits early with a correct prediction while T+A+V
would be wrong. Both are reported as percentages of all test
utterances.

\begin{table}[t]
\centering
\small
\setlength{\tabcolsep}{8pt}
\renewcommand{\arraystretch}{1.1}
\caption{Early-exit disagreements with full-modality inference on
Qwen2.5-Omni-7B. Percentages are computed over all test utterances.
Missed corrections denote incorrect early exits that T+A+V would
correct; avoided errors denote correct early exits for which
T+A+V would be wrong.}
\label{tab:early_exit_reliability}
\begin{tabular}{lcc}
\toprule
\textbf{Dataset}
& \textbf{Missed corrections}
& \textbf{Avoided errors} \\
\midrule
IEMOCAP & 4.1\% & 8.8\% \\
MELD    & 4.6\% & 7.5\% \\
MOSEI   & 2.2\% & 4.3\% \\
MOSI    & 2.0\% & 5.4\% \\
\bottomrule
\end{tabular}
\end{table}

As shown in Table~\ref{tab:early_exit_reliability}, missed
corrections account for 2.0--4.6\% of test utterances, whereas
avoided errors account for 4.3--8.8\%. On every benchmark,
correct early exits for which full-modality inference would be
wrong outnumber incorrect early exits that it would correct.
These results show that early exiting can preserve correct
predictions that would otherwise be lost when additional
modalities are introduced. 

\begin{wraptable}{r}{0.4\textwidth}
\vspace{-15pt}
\centering
\small
\setlength{\tabcolsep}{6pt}
\renewcommand{\arraystretch}{1.1}
\caption{Percentage of incorrect exits corrected by a later tier
on Qwen2.5-Omni-7B, computed over incorrect exits at each tier.
Correctness is measured against dataset labels.}
\label{tab:early_exit_correction}
\begin{tabular}{lcc}
\toprule
\textbf{Dataset} & \textbf{Exit at T} & \textbf{Exit at T+A} \\
\midrule
IEMOCAP & 2.3\% & 10.3\% \\
MELD    & 2.5\% & 8.6\% \\
MOSEI   & 1.2\% & 7.9\% \\
MOSI    & 4.0\% & 9.5\% \\
\bottomrule
\end{tabular}
\vspace{-10pt}
\end{wraptable}

\paragraph{Correction potential by exit tier.}
We further examine the proportion of incorrect early exits that a
later tier would have corrected, separately for T and T+A, using
Qwen2.5-Omni-7B. Each percentage in
Table~\ref{tab:early_exit_correction} is computed over all incorrect
predictions exiting at the corresponding tier, with correctness
measured against the dataset labels. Only 1.2--4.5\% of incorrect
exits at T and 7.9--10.3\% at T+A would have been corrected by a
later tier. Thus, most errors among early exits persist under the
evaluated later-tier predictions.

\paragraph{Empty prediction sets.}
We examine empty prediction sets on Qwen2.5-Omni-7B using the CSC-selected budgets. Table~\ref{tab:empty_sets} reports their frequency among samples reaching each tier. Empty sets do not occur at T and affect 3.2--14.2\% of samples reaching T+A. We further evaluate forwarding these samples to T+A+V while keeping the budgets unchanged. Forwarding changes wF1 by at most 0.6 points while reducing LRR by 3.0--14.3 percentage points and lowering time utility on all four datasets. These results support exiting on empty sets as an efficiency choice. 

\begin{table}[t]
\centering
\small
\setlength{\tabcolsep}{8pt}
\renewcommand{\arraystretch}{1.1}
\caption{Empty-set frequencies and the effect of forwarding empty-set samples on Qwen2.5-Omni-7B. Frequencies are computed over samples reaching each tier. Changes are Forward minus Exit with the same budgets; LRR differences are in percentage points (pp).}
\label{tab:empty_sets}
\begin{tabular}{lccrrr}
\toprule
\textbf{Dataset} & \textbf{Empty at T} & \textbf{Empty at T+A}
& $\Delta$ wF1 & $\Delta$ LRR (pp) & $\Delta U_t$ \\
\midrule
IEMOCAP & 0.0\% & 5.1\%  & $-0.5$ & $-4.9$  & $-9.7$ \\
MELD    & 0.0\% & 14.2\% & $+0.2$ & $-14.3$ & $-10.1$ \\
MOSEI   & 0.0\% & 3.2\%  & $+0.1$ & $-3.0$  & $-6.9$ \\
MOSI    & 0.0\% & 4.7\%  & $-0.6$ & $-4.6$  & $-8.0$ \\
\bottomrule
\end{tabular}
\end{table}

\subsection{Qualitative Failure Cases}
\label{sec:failure_cases2}
\begin{figure*}[t]
\begin{ReasoningBox}[title=Case Study C, breakable=false]
\textbf{Question:}
Given the video frames from the conversation, the audio clip of the utterance, and the text transcript: \textit{"Oh, my baby!"}
Classify the emotion into exactly one of the following: \\
A. anger \quad B. disgust \quad C. fear \quad D. joy \quad E. neutral \quad F. sadness \quad G. surprise

\textbf{Correct Answer:} \textcolor{correctgreen}{sadness}

\vspace{6pt}
\begin{minipage}[t]{0.58\textwidth}
\vspace{0pt}
\textbf{Model Prediction:}\\[2pt]
\textbf{Tier 1 (T):}\quad joy: $\mathbf{0.995}$, surprise: $0.005$, sadness: $0.000$, disgust: $0.000$, fear: $0.000$, anger: $0.000$, neutral: $0.000$\\
Prediction: \textcolor{wrongred}{joy}

\vspace{4pt}
{\color{gray!70}
\textbf{Tier 2 (T+A):}\quad joy: $\mathbf{0.948}$, sadness: $0.029$, surprise: $0.022$, disgust: $0.000$, fear: $0.000$, neutral: $0.000$, anger: $0.000$\\
Prediction: joy}

\vspace{4pt}
{\color{gray!70}
\textbf{Tier 3 (T+A+V):}\quad sadness: $\mathbf{1.000}$, surprise: $0.000$, anger: $0.000$, joy: $0.000$, neutral: $0.000$, disgust: $0.000$, fear: $0.000$\\
Prediction: sadness}
\end{minipage}%
\hfill
\begin{minipage}[t]{0.38\textwidth}
\vspace{0pt}
\centering
\includegraphics[width=\linewidth]{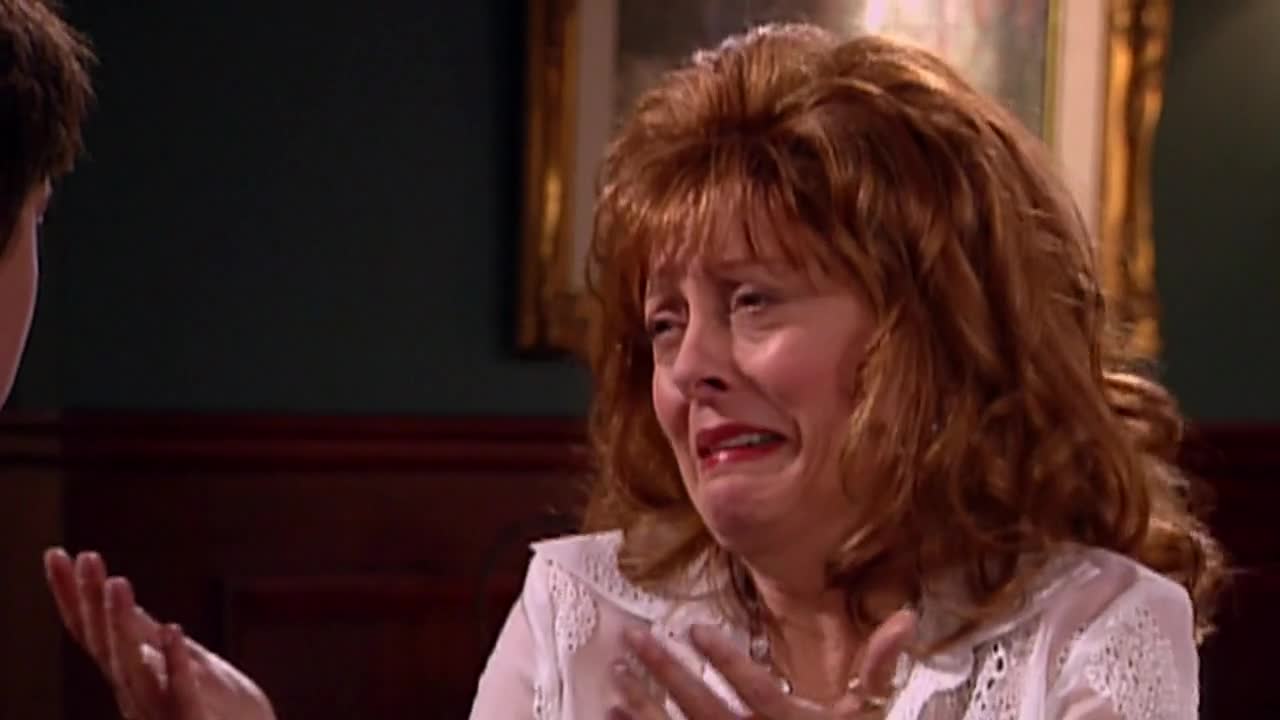}
\end{minipage}
\end{ReasoningBox}
\vspace{-10pt}
\caption{Failure case C: a premature exit at Tier 1. The text alone yields a confident but incorrect singleton, so \ours stops before the visual cue (a crying speaker) that would correct the prediction. Grey text indicates tiers that were not invoked.}
\label{fig:case-study-fail-1}
\vspace{-10pt}
\end{figure*}

\begin{figure*}[t]
\begin{ReasoningBox}[title=Case Study D, breakable=false]
\textbf{Question:}
Given the video frames from the conversation, the audio clip of the utterance, and the text transcript: \textit{"I guess this is why I came."}
Classify the emotion into exactly one of the following: \\
A. angry \quad B. excited \quad C. frustrated \quad D. happy \quad E. neutral \quad F. sad

\textbf{Correct Answer:} \textcolor{correctgreen}{happy}

\vspace{6pt}
\begin{minipage}[t]{0.58\textwidth}
\vspace{0pt}
\textbf{Model Prediction:}\\[2pt]
\textbf{Tier 1 (T):}\quad neutral: $\mathbf{0.820}$, happy: $0.143$, sad: $0.028$, excited: $0.008$, frustrated: $0.001$, angry: $0.000$\\
Prediction: \textcolor{wrongred}{neutral}

\vspace{4pt}
\textbf{Tier 2 (T+A):}\quad sad: $\mathbf{0.998}$, neutral: $0.002$, frustrated: $0.000$, excited: $0.000$, angry: $0.000$, happy: $0.000$\\
Prediction: \textcolor{wrongred}{sad}

\vspace{4pt}
{\color{gray!70}
\textbf{Tier 3 (T+A+V):}\quad happy: $\mathbf{0.999}$, neutral: $0.001$, excited: $0.000$, sad: $0.000$, angry: $0.000$, frustrated: $0.000$\\
Prediction: happy}
\end{minipage}%
\hfill
\begin{minipage}[t]{0.38\textwidth}
\vspace{0pt}
\centering
\includegraphics[width=\linewidth]{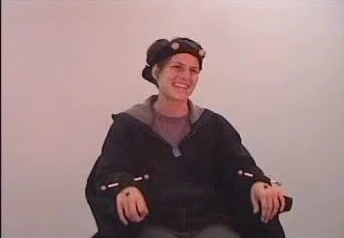}
\end{minipage}
\end{ReasoningBox}
\vspace{-10pt}
\caption{Failure case D: a premature exit at Tier 2. The text is ambiguous, and the acoustic cues drive the audio-augmented prediction to a confident but incorrect sad; the speaker's smile, visible only at Tier 3, would have corrected it. Grey text indicates the tier that was not invoked.}
\label{fig:case-study-fail-2}
\vspace{-10pt}
\end{figure*}

To complement the successful examples in the Case Study section, Figures~\ref{fig:case-study-fail-1} and~\ref{fig:case-study-fail-2} show two typical failure modes of \ours{} on Qwen3-Omni-30B, where the cascade exits early with an incorrect prediction although full-modality inference would be correct, as measured against the dataset labels. In the first example, the line ``Oh, my baby!'' reads as an affectionate exclamation, so the text tier assigns almost all probability mass to joy and forms a confident singleton; \ours{} therefore stops at Tier 1, while incorporating video of the crying speaker at Tier 3 yields sadness, matching the annotation. In the second example, the text is ambiguous and is predicted as neutral at Tier 1, so \ours{} escalates to Tier 2; however, the audio-augmented prediction collapses to a confident but incorrect sad, and incorporating the speaker's smile at Tier 3 changes the prediction to the annotated label happy. Both cases illustrate that a singleton prediction set only reflects uncertainty given the observed modalities, so \ours{} can exit before additional cues yield a prediction matching the dataset label. Such agreement does not establish which cue reflects the utterance's true emotion.

\begin{figure*}[t]
\begin{ReasoningBox}[title=Case Study A, breakable=false]
\textbf{Question:}
Given the video frames from the conversation, the audio clip of the utterance, and the text transcript: \textit{"Yes. Don't you know that? I mean, you're probably the one you wishes you were with somebody else, somebody who didn't take everything so hard and who knew how to enjoy herself."}
Classify the emotion into exactly one of the following:  \\
A. angry \quad B. excited \quad C. frustrated \quad D. happy \quad E. neutral \quad F. sad

\textbf{Correct Answer:} \textcolor{correctgreen}{sad}

\vspace{6pt}
\begin{minipage}[t]{0.58\textwidth}
\vspace{0pt}
\textbf{Model Prediction:}\\[2pt]
\textbf{Tier 1 (T):}\quad frustrated: $\mathbf{0.525}$, sad: $0.464$, angry: $0.009$, neutral: $0.002$, excited: $0.000$, happy: $0.000$\\
Prediction: \textcolor{wrongred}{frustrated}

\vspace{4pt}
\textbf{Tier 2 (T+A):}\quad sad: $\mathbf{0.969}$, frustrated: $0.029$, neutral: $0.001$, angry: $0.000$, excited: $0.000$, happy: $0.000$\\
Prediction: \textcolor{correctgreen}{sad}

\vspace{4pt}
\vspace{4pt}
{\color{gray!70}
\textbf{Tier 3 (T+A+V):}\quad 
 sad: $\mathbf{0.980}$, neutral: $0.012$, frustrated: $0.008$, angry: $0.000$, excited: $0.000$, happy: $0.000$}\\
Prediction: \textcolor{correctgreen}{sad}
\end{minipage}%
\hfill
\begin{minipage}[t]{0.38\textwidth}
\vspace{0pt}
\centering
\includegraphics[width=\linewidth]{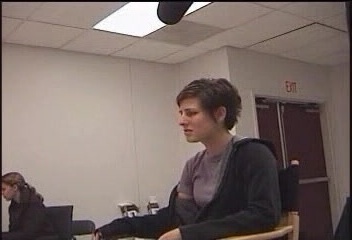}
\end{minipage}
\end{ReasoningBox}
\vspace{-10pt}
\caption{Case study A where the audio modality substantially reduces uncertainty. Grey text indicates a tier that was not invoked.}
\label{fig:case-study-2}
\vspace{-10pt}
\end{figure*}

\begin{figure*}[t]
\begin{ReasoningBox}[title=Case Study B, breakable=false]
\textbf{Question:}
Given the video frames from the conversation, the audio clip of the utterance, and the text transcript: \textit{"Uh-huh! Why do you think it takes me so long to answer the door?"}
Classify the emotion into exactly one of the following: \\
A. anger \quad B. disgust \quad C. fear \quad D. joy \quad E. neutral \quad F. sadness \quad G. surprise

\textbf{Correct Answer:} \textcolor{correctgreen}{joy}

\vspace{6pt}
\begin{minipage}[t]{0.58\textwidth}
\vspace{0pt}
\textbf{Model Prediction:}\\[2pt]
\textbf{Tier 1 (T):}\quad surprise: $\mathbf{0.450}$, anger: $0.309$, neutral: $0.114$, joy: $0.089$, fear: $0.029$, disgust: $0.006$, sadness: $0.003$\\
Prediction: \textcolor{wrongred}{surprise}

\vspace{4pt}
\textbf{Tier 2 (T+A):}\quad anger: $\mathbf{0.664}$, joy: $0.277$, neutral: $0.029$, surprise: $0.018$, disgust: $0.006$, sadness: $0.004$, fear: $0.002$\\
Prediction: \textcolor{wrongred}{anger}

\vspace{4pt}
\textbf{Tier 3 (T+A+V):}\quad joy: $\mathbf{0.977}$, neutral: $0.014$, surprise: $0.005$, anger: $0.002$, disgust: $0.001$, fear: $0.001$, sadness: $0.000$\\
Prediction: \textcolor{correctgreen}{joy}
\end{minipage}%
\hfill
\begin{minipage}[t]{0.38\textwidth}
\vspace{0pt}
\centering
\includegraphics[width=\linewidth]{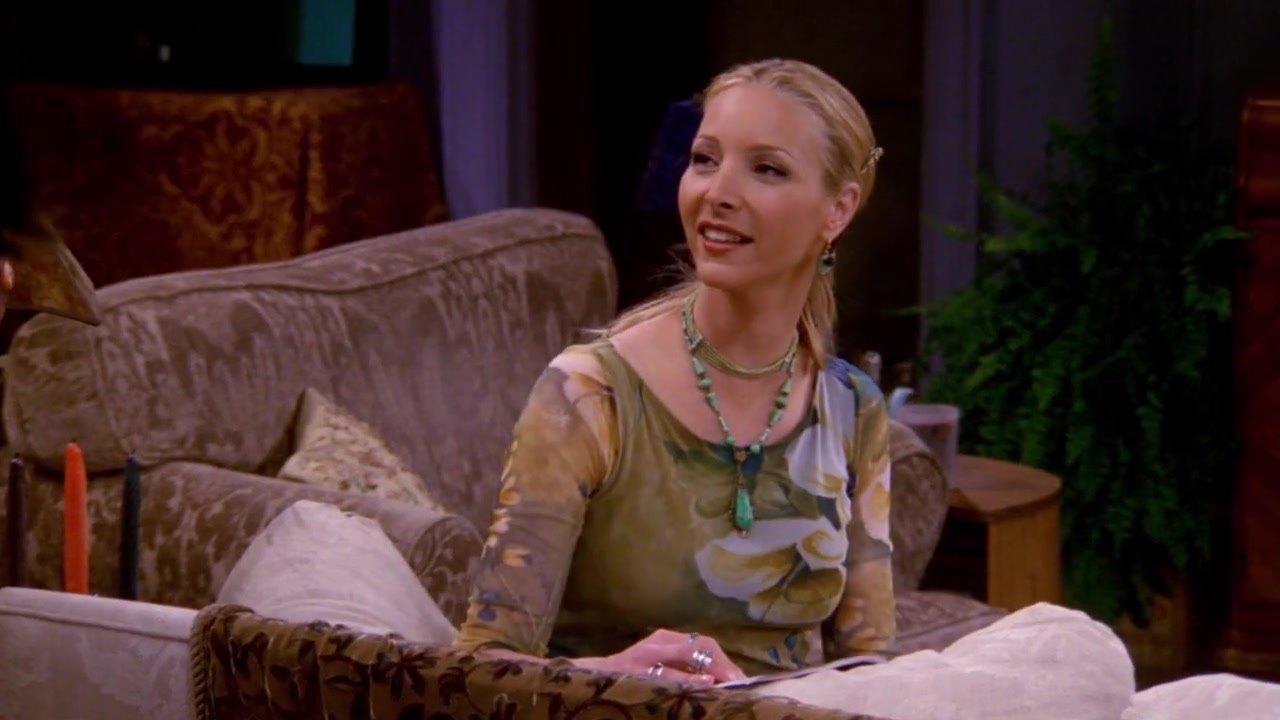}
\end{minipage}
\end{ReasoningBox}
\vspace{-10pt}
\caption{Case study B, where both the text and audio modalities are insufficient to determine the emotion.}
\label{fig:case-study}
\end{figure*}
\section{Case Study}
To better understand how \ours works with individual statements, we look at two examples in Figures \ref{fig:case-study-2} and \ref{fig:case-study}. The first example shows text that can be interpreted as either frustrated or sad. The model would choose the wrong label if it only considered Tier 1. However, when the speaker's slow, downcast delivery is included at Tier 2, the prediction narrows down to one label, allowing the process to finish without analyzing the visual cues. The second example is more complicated. The line "Why do you think it takes me so long to answer the door?" sounds like a complaint in text form. The animated delivery actually leads the audio-augmented prediction toward anger. Only when we add the visual cue (a relaxed, smiling speaker) at Tier 3 does the model correctly classify the label as joy. These examples illustrate how conformal set size can defer ambiguous predictions and enable correction at later tiers, although confident errors and unhelpful escalation remain possible.

\end{document}